%% file: templateArxiv.tex
\documentclass{article}

\usepackage{PRIMEarxiv}

\usepackage[utf8]{inputenc} % allow utf-8 input
\usepackage[T1]{fontenc}    % use 8-bit T1 fonts
\usepackage{hyperref}       % hyperlinks
\usepackage{url}            % simple URL typesetting
\usepackage{booktabs}       % professional-quality tables
\usepackage{amsfonts}       % blackboard math symbols
\usepackage{nicefrac}       % compact symbols for 1/2, etc.
\usepackage{microtype}      % microtypography
\usepackage{lipsum}
\usepackage{fancyhdr}       % header
\usepackage{graphicx}       % graphics
\graphicspath{{media/}}     % organize your images and other figures under media/ folder
    
\usepackage{tabulary}
\usepackage{array}
\usepackage{colortbl}

\usepackage{color,soul}
\title{SHARP: An Adaptable, Energy-Efficient Accelerator for Recurrent Neural Networks
}

\author{
  Reza Yazdani, Olatunji Ruwase, Minjia Zhang, Yuxiong He \\
  Microsoft \\
  \texttt{\{reyazda, olruwase, minjiaz, yuxhe\}@microsoft.com} \\
   \And
  Jose-Maria Arnau, Antonio Gonz\'alez \\
  Universitat Politecnica de Catalunya \\
  Barcelona, Spain\\
  \texttt{\{jarnau, antonio\}@ac.upc.edu} \\
}

\begin{document}
\maketitle


\begin{abstract}
 The effectiveness of Recurrent Neural Networks (RNNs) for tasks such as Automatic Speech Recognition has fostered interest in RNN inference acceleration. Due to the recurrent nature and data dependencies of RNN computations, 
prior work has designed customized architectures specifically tailored to the computation pattern of RNN, getting high computation efficiency for certain chosen model sizes. However, given that the dimensionality of RNNs varies a lot for different tasks, it is crucial to generalize this efficiency to diverse configurations. 

In this work, we identify adaptiveness as a key feature that is missing from today's RNN accelerators. In particular, we first show the problem of low resource-utilization and low adaptiveness for the state-of-the-art RNN
implementations on GPU, FPGA and ASIC architectures. To solve these issues, we propose an intelligent 
tiled-based dispatching mechanism for increasing the adaptiveness of RNN computation, 
in order to efficiently handle the data dependencies. 
To do so, we propose Sharp as a hardware accelerator, 
which pipelines RNN computation using an effective scheduling scheme to hide most of 
the dependent serialization.
Furthermore, Sharp employs dynamic reconfigurable architecture to adapt to the model's characteristics. 

Sharp achieves 2x, 2.8x, and 82x speedups on average, considering different RNN models and resource budgets, 
compared to the state-of-the-art ASIC, FPGA, and GPU implementations, respectively. Furthermore, we provide 
significant energy-reduction with respect to the previous solutions, due to the low power dissipation of Sharp 
(321 GFLOPS/Watt).
\end{abstract}


\maketitle

\sloppy

%----------------------------------------------------------------------------------------
\input{introduction}

%----------------------------------------------------------------------------------------

\input{background}
\input{lstmDesign}

%----------------------------------------------------------------------------------------
\input{Scheduling}

%----------------------------------------------------------------------------------------
\input{reconfigurability}

%----------------------------------------------------------------------------------------
\input{methodology}

%----------------------------------------------------------------------------------------
\input{results}

%----------------------------------------------------------------------------------------
\input{related_work}

%----------------------------------------------------------------------------------------
\input{conclusion}

\input{acknowledgement}

%Bibliography
\bibliographystyle{unsrt}  
\bibliography{references}

\end{document}

%% file: introduction.tex
\section{Introduction}\label{s:Introduction}

Recurrent Neural Networks (RNN) represent a well-known Deep Learning (DL) 
model~\cite{doi:10.1162/neco.1997.9.8.1735, DBLP:journals/corr/ChungGCB14, DBLP:journals/corr/abs-1801-01078},
with increasing popularity for applications that are based on sequence-to-sequence processing
~\cite{DBLP:journals/corr/SutskeverVL14, DBLP:journals/corr/VinyalsTBE14, DBLP:journals/corr/DonahueHGRVSD14, DBLP:journals/corr/VenugopalanRDMD15},
such as speech recognition~\cite{DBLP:journals/corr/MiaoGM15} and
machine translation~\cite{DBLP:journals/corr/ChoMGBSB14}. 
A key attribute of this class of neural networks is 
that they use past information to improve model accuracy. 
Long-Short-Term-Memory (LSTM)~\cite{doi:10.1162/neco.1997.9.8.1735} and Gated-Recurrent Unit (GRU) are the two most commonly used 
RNN. They can potentially remember useful information over a long period of time, providing high accuracy. 
RNNs have shown great effectiveness in many sequence processing problems and have fostered state-of-the-art research innovations, 
such as in natural language processing tasks, e.g., machine reading 
comprehension~\cite{DBLP:journals/corr/SeoKFH16, DBLP:journals/corr/abs-1906-06045, Wang2018AnLM} and 
language modeling~\cite{unknown,DBLP:journals/corr/abs-1903-07435, DBLP:journals/corr/abs-1902-06704}, 
and speech recognition~\cite{Park_2019, 8683660}.
\begin{figure}[t!]
\centering
\includegraphics[width=3.345in]{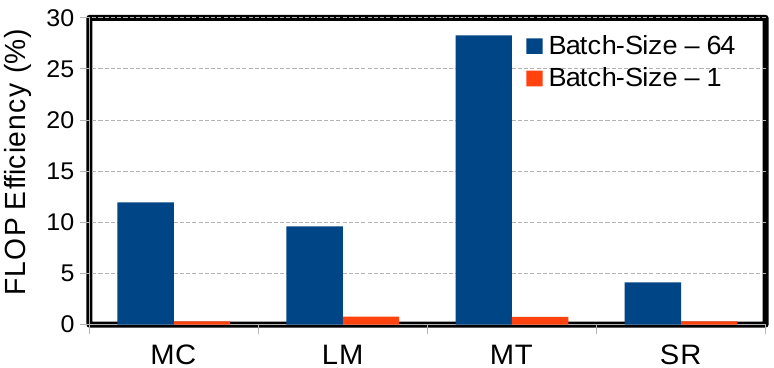}
\vskip -0.5em
\caption{Flop-efficiency of Titan V GPU, performing 
different real-world sequence processing applications, i.e. Machine Comprehension (MC)~\cite{DBLP:journals/corr/SeoKFH16}, 
Speech Recognition (SR)~\cite{DBLP:journals/corr/MiaoGM15}, Language Modeling (LM)~\cite{DBLP:journals/corr/ZarembaSV14}, and
Machine Translation (MT)~\cite{DBLP:journals/corr/WuSCLNMKCGMKSJL16}
using cuDNN library~\cite{DBLP:journals/corr/ChetlurWVCTCS14} and enabling TCUs with mixed-precision. }
%We consider 50 recurrent time-steps for all the cases, except for the SR that we use 370.}
\label{f:flopEfficiency}
\end{figure}

%The ability to keep track of past history comes with intrinsic challenges 
%for LSTM computation, such as relying on many recurrent steps in order to obtain an 
%accurate inference mechanism. This results in lots of dependencies 
%and serialization, limiting the amount of parallelism exploited by
%multicore CPUs or GPUs~\cite{DBLP:journals/corr/abs-1711-07480, cuDnn}. Figure~\ref{f:flopEfficiency} shows
%the FLOP efficiency, i.e. the relative FLOPs performance to the the peak, for a high-end 
%GPU (Titan V), when running different applications using the most recent cuDNN 
%library~\cite{DBLP:journals/corr/ChetlurWVCTCS14}. 
To meet the requirements of real-time inference at large scale, a high-performance 
and energy efficient accelerator for RNN is highly desired. However, two reasons 
make it very difficult to accomplish efficient RNN computation by CPUs or GPUs 
in parallel~\cite{DBLP:journals/corr/abs-1711-07480, cuDnn}: (1) recurrent behaviour
of RNN architecture which imposes several data-dependencies, (2) limited parallel tasks
due to the enforced low batch size by Service-Level Agreements (SLAs) in the inference
evaluation~\cite{Holmes:2019:GLS:3302424.3303949,8416814}.

First, RNN has complex dependencies and serialization, which limits the amount of parallelism that can be exploited by many cores. Take the state-of-the-art RNN in cuDNN library~\cite{DBLP:journals/corr/ChetlurWVCTCS14} on GPU as an example, Figure~\ref{f:flopEfficiency} shows that FLOP efficiency, i.e. the relative FLOPs performance to the peak, for a high-end GPU (Titan V), when running
different applications.
Note that the evaluation is measured by
enabling tensor-core-units (TCUs) using the FP16/FP32 mixed-precision, as explained in~\cite{NVIDIATCU}. As seen, GPU is
extremely under-utilized when performing services of batch size 1. Furthermore, 
even when using larger batch size of 64, the GPU achieves
moderate utilization, between 4\% to 28\% of peak performance.
The reason is that GPU only operates efficiently when there is high
level of parallelism available, such as for training. However, for RNN inference,
even though the amount of computation (matrix-vector multiplications) increases for 
long sequences, as in speech recognition or machine translation tasks, 
the parallelism is limited due to recurrent steps and data dependencies.

%Furthermore, the inference engines are commonly required to operate on small batch sizes of a few requests, at a time, 
%in order to meet the Service-Level Agreements (SLAs), or the safety requirements~\cite{Holmes:2019:GLS:3302424.3303949,8416814}.
Second, for the online inference scenario, queries come in one-by-one and have stringent latency SLA, often in single milliseconds~\cite{Holmes:2019:GLS:3302424.3303949,8416814}. 
This requirement further reduces data reuse and available parallelism in RNN inference. 
Recently, there have been several efforts on either CPU~\cite{216077}, or GPU~\cite{Holmes:2019:GLS:3302424.3303949, cuDnn}, to improve the efficiency of RNN inference. However, they show poor scalibility
for either small or large models with different sequence length. 
%For instance, we have evaluated GRNN~\cite{Holmes:2019:GLS:3302424.3303949}, 
%a scalable implementation on GPU, for various hidden time-steps required for some 
%applications such as speech recognition, which shows poor performance for large sequences.

%\begin{figure}[t!]
%\centering
%\vskip -1.5em
%\includegraphics[width=2.8in, height = 1.4in]{figures/LSTMarch.pdf}
%\caption{LSTM's structure overview.}
%\label{f:lstmStruct}
%\vskip -1.5em
%\end{figure}

Because the performance of RNN on general-purpose processors has difficulties to meet 
the requirements of real-time inference in modern applications, 
accelerating RNN through either customized
architectures~\cite{DBLP:journals/corr/abs-1711-07480, DBLP:journals/corr/abs-1803-06305} or 
neural processing units (NPU)~\cite{8416814, DBLP:journals/corr/JouppiYPPABBBBB17} has been 
recently explored.
These systems are implemented on either ASICs or FPGAs. FPGAs are attractive 
for their cost and reconfigurability, whereas ASICs are more energy-efficient. 
%In this work, we focus on ASIC implementations. We address most of the issues 
%and challenges observed due to the variety of LSTM models, improving scalability 
%through an efficient scheduling scheme together with a reconfigurable architecture.
%Low latency and efficiency are two important metrics in 
%LSTMs which increase user satisfaction~\cite{41217}. When handling
%large-scale services such as web search, advertisement, and 
%conversational bots, LSTMs require lots of parallelism to deliver 
%a fast response~\cite{216077}. However available parallelism is limited by the recurrent nature of LSTMs, making LSTMs the most challenging type of neural networks to accelerate.
However, we show that even though previous accelerators have achieved good performance 
improvement over CPUs and GPUs, they suffer from two important issues, which 
are \emph{low resource utilization and suboptimal energy consumption}. 
There are two reasons on why existing approaches have low resource utilization: (1) they are efficient
only for certain configurations, but they become inefficient
when the model configurations start to change; (2) they do not
achieve a well-balanced execution pipeline as the available hardware
resources increase. Although the two state-of-the-art implementations, Microsoft's BrainWave~\cite{8416814}
and Google's TPU~\cite{DBLP:journals/corr/JouppiYPPABBBBB17}, 
can achieve high utilization on certain models (e.g. CNNs) and configurations, 
they only achieve an average utilization of 18\% and 3.5\% for LSTMs, respectively. 
%achieveaverage utilization of 18\% and 3.5\% for LSTMs, respectively. 
We elaborate on the scaling issues of two state-of-the-art ASIC- and FPGA-based RNN accelerators in Section~\ref{s:Challenges}.

In this paper, we propose Sharp, an adaptable and energy-efficient architecture
for RNN inference acceleration. We show that by carefully analyzing the unique challenge and special characteristics of RNN, we could combine a system-level scheduling scheme and a hardware reconfigurability design to efficiently overcome the data dependencies and low parallelism issue, which yields the
most efficient workload-dispatching configuration for each model and addresses the low resource utilization challenges from the previous designs.
In particular, we introduce the \emph{unfolded scheduling} which significantly reduces
the length of the RNN critical-path while also strictly overlapping the computations and data transfer time. Furthermore, we introduce \emph{dynamic
reconfigurability} that allows the accelerator to adaptively get high resource utilization for different RNN configurations and handle padding caused by matrixvector multiplication more effectively. Moreover, we show that even though the total
power per time unit increases marginally, our design achieves much higher energy-efficiency than existing approaches by making the RNN computations a lot faster.
% Besides, LSTM computation makes extensive use of matrix-vector operations and often incurs in some padding, 
% resulting in resource under-utilization. In our proposal, we also use reconfigurability to handle 
% padding more effectively.

\begin{figure}[t!]
\centering
\includegraphics[width=2.8in,height=1.2in]{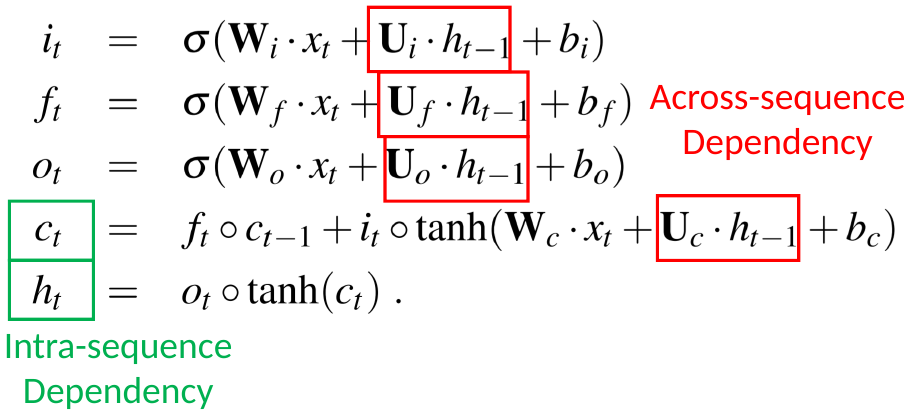}
\caption{LSTM computation overview. By looking at the formulas, we see two types of data-dependency, intra-sequence and across sequence, in an LSTM-cell.}
\label{f:lstmComp}
\end{figure}

Overall, 
%this work introduces several approaches at both system and architecture levels
%in order to optimize the design of an LSTM accelerator. 
we sum up the paper's contributions as follows:

\begin{itemize}

\item We propose Sharp, a \textbf{S}calable, \textbf{H}igh-performance RNN \textbf{A}ccelerator,
which removes pipeline stalls and resources' idleness through \textbf{R}econfiguration and better
handles \textbf{P}adding in matrix-vector multiplication.

\item We analyze RNN's critical-path delay, considering its data dependencies, and 
identify opportunities to hide the latency of sequential parts. 
We then develop a new scheduling scheme resolving all dependencies.

\item To increase the adaptability of our system running different models, 
we implement a reconfigurable compute-engine that delivers the most efficient resource
mapping.

\item %By employing the above optimization techniques in the LSTM-Sharp's pipeline,
%we can achieve 
We conduct thorough evaluation and demonstrate
average speedups of 2x, 2.8x, and 82x with respect to the state-of-the-art ASIC, FPGA, and GPU 
implementations. Furthermore, we obtain 1.7x and 7.4x higher FLOPS/Watt compared to the 
previous ASIC and FPGA designs, respectively.

%We also compare our numbers with
%the two state-of-the-art GPU implementations, cuDNN and GRNN, gaining between 
%1 to 2 orders of magnitude performance improvement.

\end{itemize} 

The remainder of this paper is organized as follows. Section~\ref{s:background}
provides some background on LSTM as an RNN network. Section~\ref{s:Challenges} furthermore introduces some 
challenges and opportunities regarding RNN acceleration design. 
Next, we introduce our proposed scheduling 
technique in Section~\ref{s:lstm_analysis}. Afterwards, we describe Sharp's design 
in Section~\ref{s:lstmDesign}. Then, Section~\ref{s:lstm_reconnfigurability} discusses the 
architecture's reconfigurability. 
We describe the evaluation methodology in Section~\ref{s:EvaluationMethodology} and show 
the experimental results in Section~\ref{s:ExperimentalResults}. 
Finally, Section~\ref{s:RelatedWork} reviews some related work, and 
Section~\ref{s:Conclusions} sums up the main conclusions of this work.

%% file: background.tex
\section{RNN Background}\label{s:background}

%LSTM networks~\cite{doi:10.1162/neco.1997.9.8.1735} can capture both short and long term 
%dependencies of an input sequence through their recurrent links. %This results in a lot of 
%%serial processing, and such inherent dependencies at LSTM computation 
%%make it the most challenging of all types of neural networks to parallelize. 
%In this section, we first go through some overview
%of LSTMs, and then elaborate more on its computation style. Finally,
%we discuss the most common scheduling schemes used in the previous LSTM
%accelerators.

RNN architectures can simply capture short term dependencies. However, exploiting long term
dependencies is challenging and useful at the same time, increasing the quality of the deep-learning
models. With this regard, Long-Short-Term-Memory (LSTM) is considered as the most successful and widely 
used RNN design, with applications in speech recognition~\cite{Park_2019, 8683660}, machine translation~\cite{DBLP:journals/corr/WuSCLNMKCGMKSJL16} and language modeling~\cite{unknown,DBLP:journals/corr/abs-1903-07435, DBLP:journals/corr/abs-1902-06704}. In this section, we
explain in detail the structure and behavior of LSTM networks.

An LSTM network is composed of a chain of LSTM cells, and each cell processes 
two vectors, $x_{t}$ and $h_{t}$ at each time step, corresponding to the input and
hidden vectors of the forward and recurrent connections, respectively. It 
employs four gates in order to recurrently update the cell state and also compute the output. 
At each recurrent phase, the gates carry out the following actions: the \textit{input}
gate ($i_{t}$) decides how the current input affects the cell-state,
while the \textit{forget} gate ($f_{t}$) removes the amount of useless information from the 
current cell-state; the \textit{cell-update} gate ($g_{t}$) modulates the amount of 
input information that is considered as candidate to update the cell-state; 
finally, the output gate ($o_{t}$) decides what information 
to emit from the cell.

\begin{figure}[t!]
\centering
\includegraphics[width=3.345in]{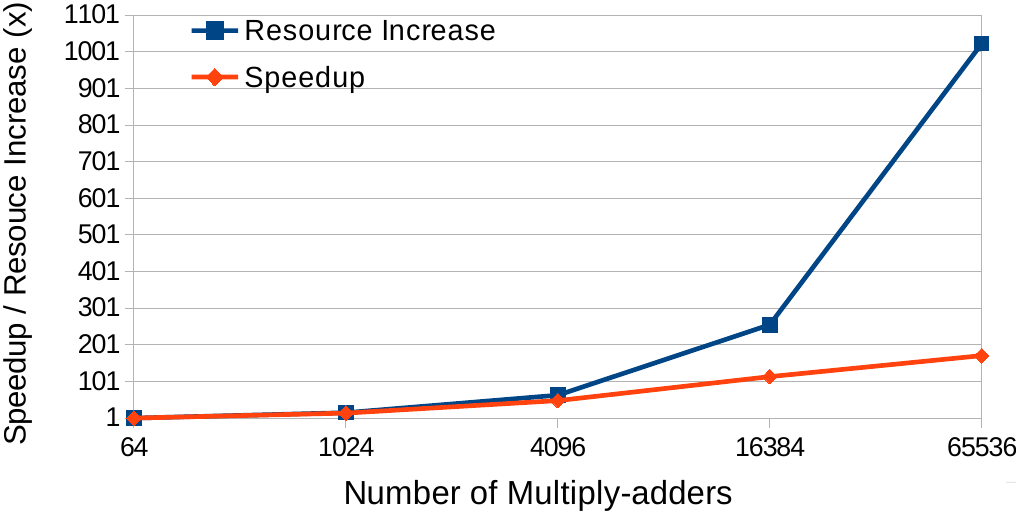}
\vskip -0.5em
\caption{BrainWave's latency and resource-utilization reported in~\cite{8416814}, for running different LSTM models.}
\vskip -1em
\label{f:brainwave_util}
\end{figure}

Figure~\ref{f:lstmComp} formulates the detailed computation of LSTM at each time step.
As depicted, each LSTM gate performs two matrix-vector multiplications (MVMs), which finally 
decide how to update the cell-state ($c_{t}$), and how to generate
the hidden output vector ($h_{t}$) that is recurrently sent to the following time step. Two kinds
of dependencies exist in these computations: intra-sequence, since
all the gate's activations must be ready before updating the cell or generating the
output; across-sequence, meaning that each step has to wait until 
its recurrent input is received from the previous time step. Later, we will discuss 
the parallelism constrains due to the sequential behavior imposed by these dependencies.

\section{Challenges and Opportunities}\label{s:Challenges}

Most of the previous accelerators are designed for a variety
of neural networks rather than being tightly optimized for RNNs, specifically LSTM. 
For instance, NPUs~\cite{8416814, DBLP:journals/corr/JouppiYPPABBBBB17} 
have the parallel multiply-accumulation (MAC) 
stage as the heart of their pipeline and are not optimized in 
case the serial part becomes the performance bottleneck for some models. On the other hand, 
customized accelerators~\cite{DBLP:journals/corr/abs-1711-07480, DBLP:journals/corr/HanKMHLLXLYWYD16} 
use a relatively small resource budget which 
therefore causes large delay for MVM, hence overlap 
the remaining LSTM computation that needs to run sequentially. 
However, when using more MACs, the issue of efficiently handling LSTM's dependencies still remains.
%In this work, we address the challenges of the acceleration design for different amount of 
%resources and model dimensions. 

Figure~\ref{f:brainwave_util} shows the latency and utilization of BrainWave
for different LSTM sizes. As the size of the hidden layers decreases, utilization
drops drastically, whereas the latency remains the same. However, an efficient
design should operate faster as the LSTM workload reduces.
The reason for such performance inefficiency is that BrainWave's pipeline is mainly 
optimized for some particular large models rather than being adaptable to various models.
{As stated in~\cite{8416814}, for small LSTM, BrainWave's utilization drops due to two main reasons:
(i) the design of large tile dimension for the multiplication units, resulting in 
wasteful work and resource under-utilization; (ii) the deep pipeline which delays the
writing of the dependent data back.} In other words, the pipeline is not well-balanced 
in assigning resources to different stages based on the models' requirements, 
which causes many stalls in case one stage is slower. 

\begin{figure}[t!]
\centering
\includegraphics[width=3.345in]{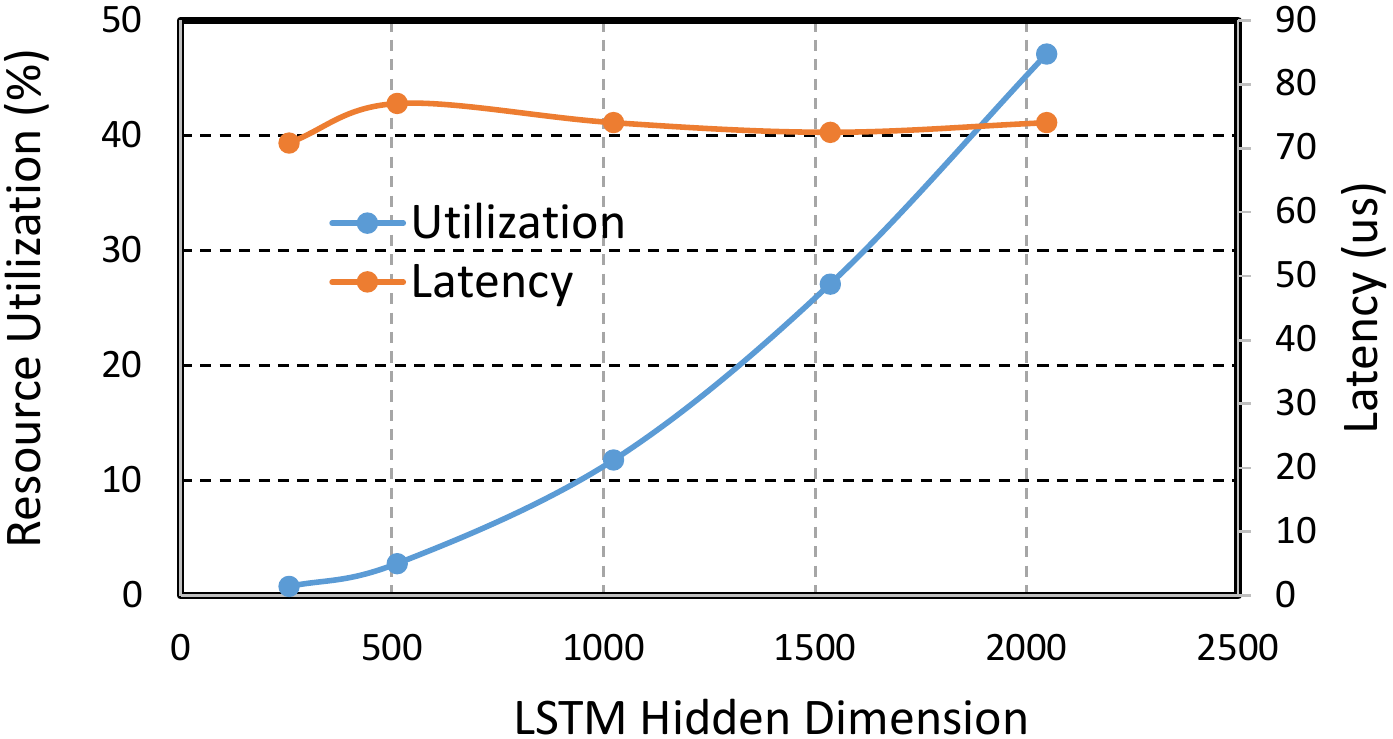}
\caption{The figure shows E-PUR's~\cite{DBLP:journals/corr/abs-1711-07480} 
speedup running EESEN~\cite{DBLP:journals/corr/MiaoGM15}, for a range of MAC
units. Due to the scalability issue, it does not achieve performance improvement
proportional to the increase in resources.}
\label{f:scalability}
\end{figure}

On the other hand, some designs obtain good performance efficiency for a specific model only when resources are limited, but are inefficient with larger resource budgets. For instance, we thoroughly evaluate E-PUR~\cite{DBLP:journals/corr/abs-1711-07480}, 
the state-of-the-art dense RNN accelerator, by experimenting across different number of multiply-adder units. Figure~\ref{f:scalability} shows
the performance improvement obtained by increasing resources when accelerating
EESEN~\cite{DBLP:journals/corr/MiaoGM15}, one of the benchmarks used 
in~\cite{DBLP:journals/corr/abs-1711-07480}. 
As seen, by raising the multiply-add units above 4K, we are not able 
to achieve an efficient speedup compared to the increase in the number of resources.

An LSTM cell computes eight different MVMs which can run in parallel at each time step. 
Previous proposals mainly use vector-vector (VV)~\cite{DBLP:journals/corr/abs-1711-07480}, vector-matrix (VM)~\cite{8416814} 
and matrix-matrix (MM)~\cite{DBLP:journals/corr/JouppiYPPABBBBB17} primitives, which are
the most simple, straightforward hardware approaches. However, they are less flexible
since their vector and matrix dimensions are set to a fixed size, which results in resources'
under-utilization in many cases. In this work, we use vector-scalar (VS) as the basic primitive and implement 
VV and VM by merging VSs in different dimensions. 
This way, we can implement resizable primitives by using VSs of different sizes.

To address the aforementioned challenges, we propose SHARP in the next section, as a 
reconfigurable RNN accelerator design with an efficient pipelining mechanism combined with a new scheduling. 

%% file: lstmDesign.tex
\section{SHARP's Architecture}\label{s:lstmDesign}

In this section, we present the architecture of SHARP.
First, we describe the accelerator's pipeline considering LSTM processing flow. 
Next, we elaborate on each pipeline stages. 
Finally, we discuss the balance between different component's 
latency in order to keep the pipeline fully utilized. 

\subsection{Overview}

\begin{figure*}[h]
\centering
\includegraphics[width=\columnwidth]{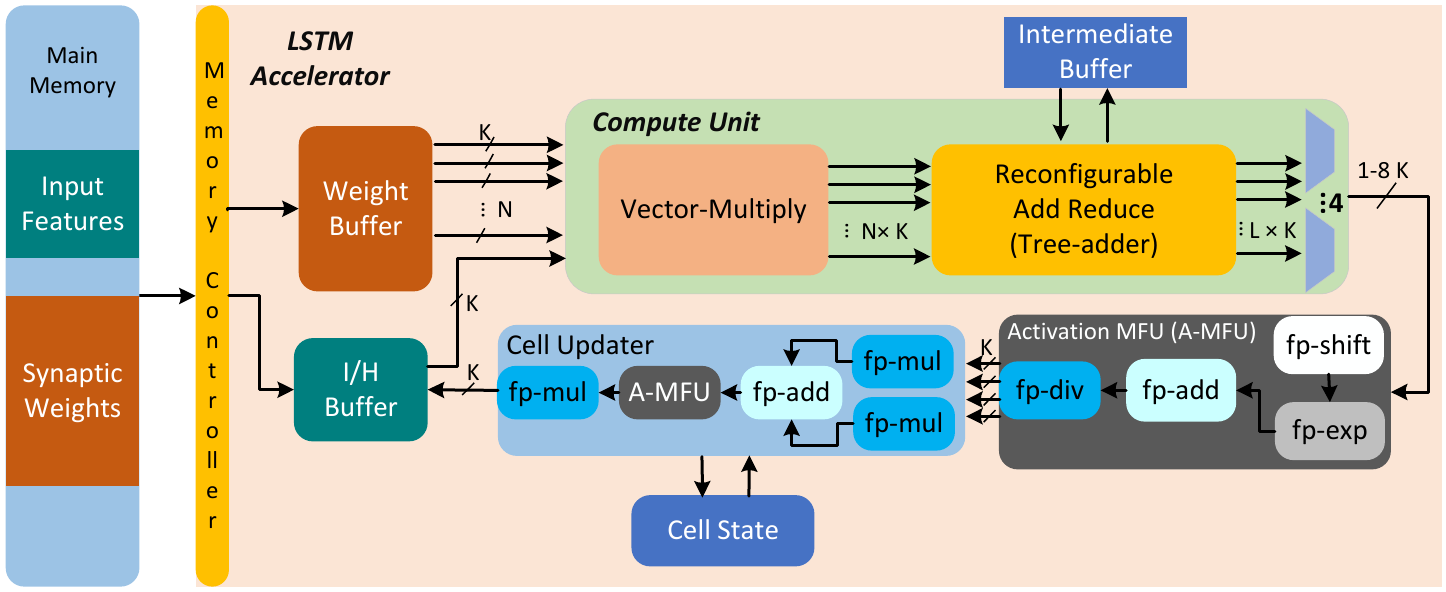}

\caption{The architecture of SHARP accelerator.}

\label{f:lstmDesign}
\end{figure*}

Figure~\ref{f:lstmDesign} illustrates the SHARP's hierarchical pipelined architecture
composed of 3 stages: \textit{Compute Unit}, \textit{Activation MFU (A-MFU)}, and 
\textit{Cell Updater}. Moreover, SHARP uses local FIFOs at all stages in order to control 
the data-flow and also decouple the producer and consumer pattern as well as computation and memory accesses.
Considering LSTM computations, the pipeline performs the following tasks at each time step: 
First, \textit{Compute Unit} multiplies the weight matrix and input/hidden vector in a tiled-based fashion,
by fetching data from weight and I/H buffers, respectively; Second, after completing each gate's matrix-vector 
multiplication (MVM) for both input and hidden vectors, \textit{A-MFU} runs the activation 
function, i.e. sigmoid or hyperbolic tangent, on the MVM's result. 
Finally, \textit{Cell Updater} uses all the four gates' activated results to update the cell-state 
and produce the hidden outputs for the next step.

In addition to the functional units, SHARP includes two memory
components in order to store the synaptic weight matrices and input and hidden
vectors necessary for one RNN layer's evaluation. This way we can avoid 
most of the  expensive off-chip memory accesses, which is identified as 
the main feature in most of the state-of-the-art hardware 
implementations~\cite{DBLP:journals/corr/abs-1711-07480, 8416814, DBLP:journals/corr/JouppiYPPABBBBB17}.
Regarding the I/H buffer, we use SRAM in order to reduce the access latency and since 
it often gets modified between sequence processing. Moreover, we model weight 
buffer as a multi-banked SRAM memory, providing terabytes per second of bandwidth in order to 
feed the MAC functional unit with maximum throughput. Due to the predictable pattern of RNN computation, we can easily
interleave the weight matrices across different memory banks and fully utilize the multiply units 
without having collisions in accessing similar memory line.
Additionally, we use two double-buffered scratchpad memories for storing the cell-state and intermediate results 
produced between the recurrent time steps. 

In the following, we go over the main difference of the SHARP's design compared to the previous work. 
Specifically, we will present the reconfigurability for the multiply-add operations and choosing the best tiling
dimension for the matrix-multiplication based on the application need and the hardware availability. 
In the next section, we discuss several scheduling methods and 
propose one which can tightly couple with SHARP's pipeline, resulting in both high throughput and low latency 
compared to the previous designs.

\subsection{Resizable MVM Tile-Engine}\label{s:lstm_mvm}

\begin{figure*}[h]
\centering
\includegraphics[width=\columnwidth]{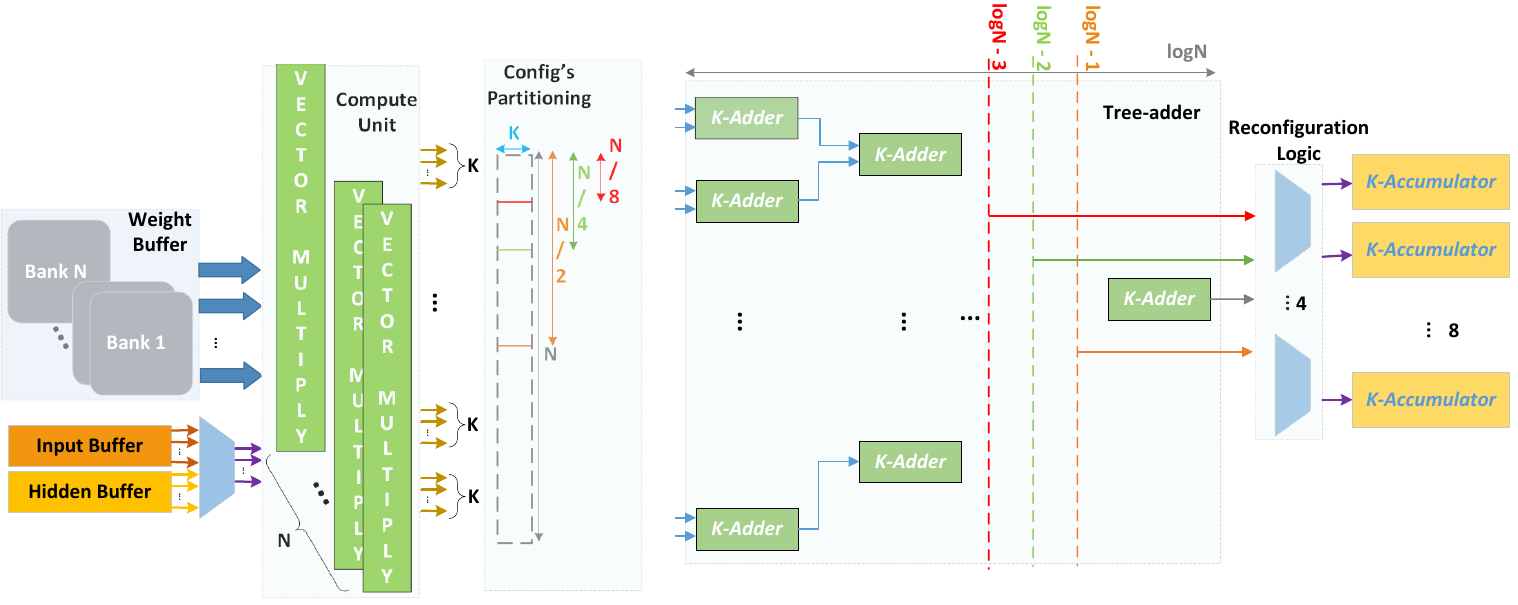}

\caption{Vector Multiply plus weight and I/H memory buffers (left) and R-Add-Reduce (right) structure. 
By using different configurations shown in figure~\ref{f:mvmConfigs}, we partition 
the multiplication outputs and reroute tree-adder's outputs using four multiplexers accordingly.
Based on the selected MVM tiling configuration, we generate 1$k$ to 8$k$ outputs.}
\label{f:recTreeAdder}
\end{figure*}

\begin{figure}[t!]
\centering
\includegraphics[width=3.345in]{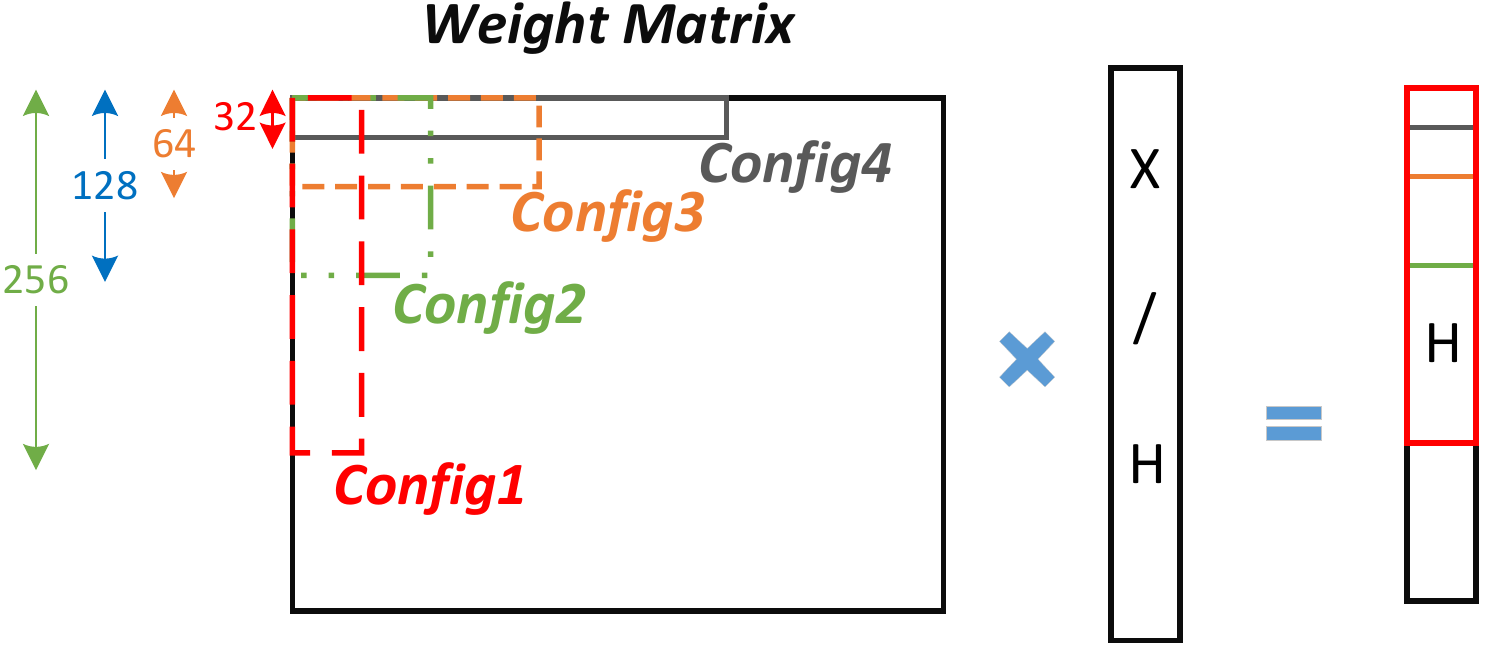}
\caption{Different MVM configurations chosen based on the $k$-width of the VS units. We consider 32 as the $k$.}
\label{f:mvmConfigs}
\end{figure}

Prior works often use the Dot Product Unit (DPU), which
operates on two vectors, to perform MVM by dispatching the weight matrix
column-wise~\cite{DBLP:journals/corr/abs-1711-07480, DBLP:journals/corr/abs-1803-06305, 8416814, DBLP:journals/corr/JouppiYPPABBBBB17}. 
However, we observe that former scheme has to reduce
the result-vectors into several outputs which may require 2
reduction levels (such as in~\cite{8416814}). In contrast, we consider row-wise selection for
the basic vector operation. 
Figure~\ref{f:recTreeAdder} (left) shows our Compute Unit structure plus
the weight and input and hidden buffers. The Compute Unit
is equipped by N K-width vector-scalar (VS) multipliers,
each multiplying an input/hidden by k-row elements of the
weight matrix and producing N K \emph{partial results}. 
The Compute Unit then generates one or multiple vectors of partial sums by accumulating the
result-vectors, requiring \emph{only 1-level reduction}. As depicted by
Figure~\ref{f:mvmConfigs}, we can allocate the N VS units both row-wise or
column-wise, generating resizable MVM tile-engines to go
over gate’s computation in a tiled fashion. In the next section,
we will show that different choices of K and VS mapping
impact performance and utilization of SHARP when
running different models.

%Figure ~\ref{f:recTreeAdder} (left) shows the \textit{Compute Unit} structure plus the weight and input and hidden buffers.
%\textit{Compute Unit} is equipped by $N$ $K$-width vector-scalar (VS)  
%multipliers, each multiplying an input/hidden by k-row elements of 
%the weight matrix and producing $N\times K$ partial results. 
%Most of the previous proposals use the Dot Product Unit (DPU)
%that operates on two vectors, by dispatching the weight matrix
%column-wise. However, we consider row-wise selection for the basic vector operation, because
%former scheme has to reduce the result-vectors into several outputs which may require 2 reduction levels (such as in~\cite{8416814}), 
%whereas ours generates one or multiple vectors of partial sums by accumulating the result-vectors, requiring 1-level reduction. 
%As depicted by Figure~\ref{f:mvmConfigs}, we can allocate the $N$ VS units either row-wise or column-wise, 
%generating resizable MVM tile-engines to go over gate's computation in 
%a tiled fashion. In the next section, we will show that 
%different choices of $K$ and VS mapping impact performance and utilization of SHARP when running different models.

We design \textit{Reconfigurable Add Reduce (R-Add-Reduce)} using a tree-adder that sums all the $K$-vector 
results into a $K$ partial sum (Figure~\ref{f:recTreeAdder}). 
This way, we reduce the results in case where all the VS units are mapped column-wise as shown in \textit{Config4} of Figure~\ref{f:mvmConfigs}. 
Therefore, \textit{R-Add-Reduce} has a maximum latency of $log(N)$, when traversing all tree-adder's levels. In order to hide 
this delay, we pipeline all the levels of tree, resulting in a 1-cycle add-reduction 
if the pipeline is full. As Figure~\ref{f:mvmConfigs} depicts, we can update between $1K$ to $8K$ accumulators by choosing the different configurations,
as we reach the 4 last levels of tree. Upon completion of each MVM, 
the accumulators are released for the next phase of the different types of RNN cell's computation.
As the computation of Figure~\ref{f:lstmComp} formulates, the input and hidden vectors must be processed
before sending out the result from \textit{R-Add-Reduce} to 
\textit{Activation MFU}.

\subsection{Gate Activation and Cell-Update}

\textit{Activation MFU} is as a configurable multi-functional unit,
composed of different arithmetic functions operating several floating-point operations
including shift, addition, division, and exponentiation. By combining these units,
we implement the two activation functions (sigmoid and hyperbolic tangent) 
applied to the gate's outputs. We use the same approach proposed in~\cite{DBLP:journals/corr/abs-1711-07480}, in order
to configure MFU data transfer based on each activation function. For instance, MFU carries out
the following actions to get the sigmoid of $X$:

\begin{equation} \label{eq1}
    X = e^X  \longrightarrow  X = X + 1  \longrightarrow X = \frac{1}{X}
\end{equation}
Based on our synthesis evaluation using Synopsys Design Compiler~\cite{s111} combined with 32nm
technology library, we calculate MFU's critical-path-delay as 29.14 ns for hyperbolic tangent function. 
We then partition these operations as shown in Figure~\ref{f:lstmDesign}, to efficiently pipeline them,
achieving 1-cycle latency for performing the activation function on each gate's output.

As soon as all the four gates' activation results are ready, \textit{Cell Updater} starts 
the following two sequential tasks: updating the cell-state, and producing the hidden 
outputs. Regarding the calculation of cell-state ($c_t$), \textit{Cell Updater} uses the 
outputs of input, forget and cell-update gates, plus the previous cell-state 
(see Figure~\ref{f:lstmComp}). Then, to compute the hidden outputs ($h_t$), a hyperbolic 
tangent is applied to the new cell state, and the result is multiplied by the mask generated 
from the output gate. Therefore, \textit{Cell-Updater} also includes an A-MFU plus several point-wise 
fp16-multiply vector units and a fp32-add vector unit. We pipeline all the operations in order to assure that the calculation of every $\frac{K}{4}$ elements of hidden
outputs (combining the 4 gates outputs) finish at each cycle (providing that pipeline is always full).

\subsection{Pipeline Efficiency}\label{s:pipeline_efficiency}

RNN computation consists of several data dependencies, which makes it a lot challenging 
to design an efficient pipeline to overlap the independent and the sequential parts.
We employ \textit{Unfolded} scheduler (Section~\ref{s:lstm_analysis}) on top of 
our pipeline design in order to improve performance under different resource configurations.
As the MVM operations include an important share of RNN processing,
the main focus of previous proposals is to provide more 
parallelism by increasing MAC resources and reduce RNN latency. 
However, we observe that for several applications RNN computation can cause a lot of stalls due 
to its serialization, which limits parallelization. For instance, LSTM models with small
dimensions that process long sequences are the most tangible examples that require dealing with
lots of dependencies besides the parallel task of MVMs. Thus, we cannot achieve a reasonable 
performance improvement by only providing high amount of parallelism. %due to the large degree of 
%resource under-utilization and idleness.

In order to achieve the best throughput, all the pipeline stages should be kept fully utilized. This means 
that the different components must have similar latency in order for the pipeline to flow without 
any stalling. Otherwise, it happens that one stage operates faster or slower than others,
causing stalls or idleness. This results in under-utilization because of uneven distribution of the amount 
of work and the number of resources between different stages. In our design, we divide the
workload of RNN based on their types of operation, whether they are dependent or independent, and then 
explore various number of resources for each part. Our experiments show that there is a high correlation 
between the scheduling scheme and the way pipeline resources are allocated to each part of RNN computation. 
Furthermore, there is not just one best resource mapping (tiling dimension) to 
evaluate all the RNN models, since each model has different requirements based on the ratio
between parallel and serial tasks. Therefore, by adding some level of reconfigurability at the \textit{Compute Unit} 
and \textit{R-Add-Reduce} components, we increase our design's adaptability by tailoring the best 
configuration for each model. We elaborate on the reconfigurability technique in Section~\ref{s:lstm_reconnfigurability}.

 \begin{figure*}[h]
\centering
\vskip -1.3em
\includegraphics[width=\columnwidth]{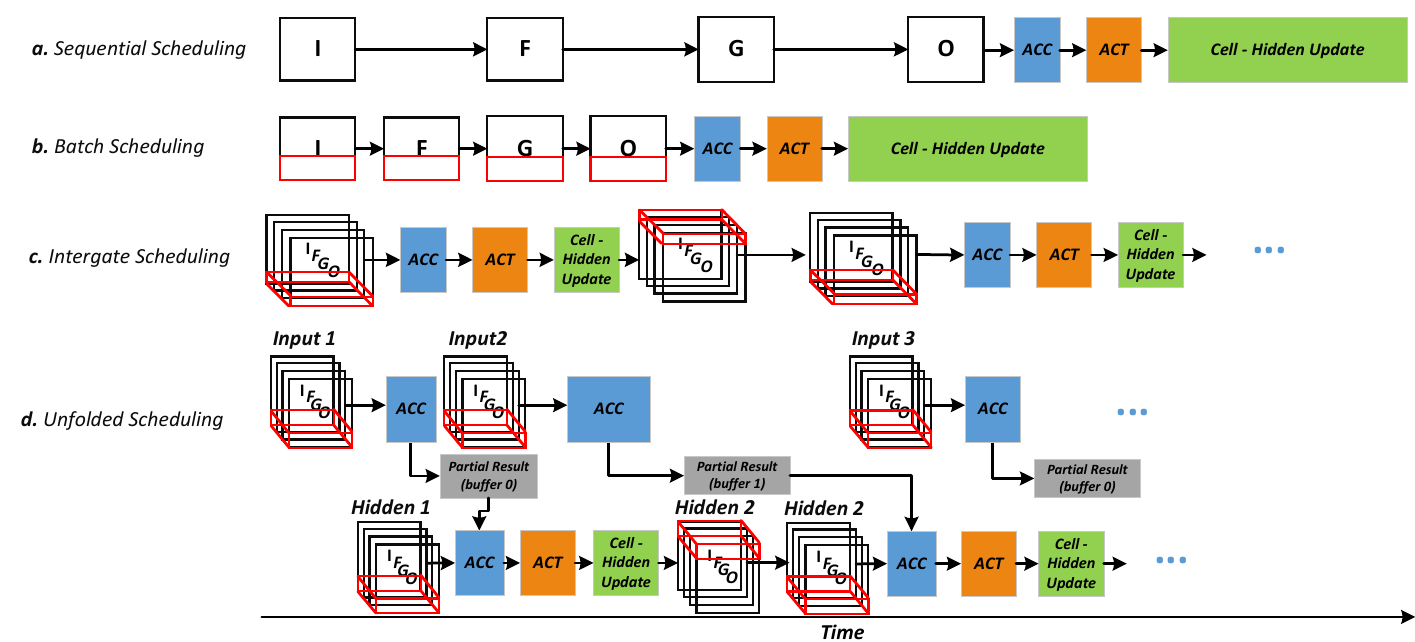}

\caption{The time-line of LSTM critical-path computation using the four scheduling schemes: a. 
\textit{Sequential}, b. \textit{Batch}, c. \textit{Intergate}, and d. \textit{Unfolded}. 
The $I$, $F$, $G$, and $O$ squares represents the MVM of the \textit{Input}, \textit{Forget},
\textit{Cell-update}, and \textit{Output} gates, respectively. Red boxes show the matrix batches dispatched to
the MVM tile-engine.
Long and short arrows show the completion of
the whole and a batch of the gate(s), respectively. Moreover, \textit{ACC} and \textit{ACT} stands for Accumulate and Activation. We 
hide the detail for the accumulation and activation of the intermediate gates, since they
are overlapped by the MVM tile operations.  
In order to have equal resources for all the schedules, \textit{Intergate} and \textit{Unfolded} 
techniques take a batch size (for each gate) four and two times smaller than {Batch} scheduling, 
respectively.}
\label{f:unfoldedMotivation}
\end{figure*}

%% file: Scheduling.tex
\section{SHARP's LSTM Schedule}\label{s:lstm_analysis}
%\begin{figure}[t!]
%\centering
%\includegraphics[width=2.8in,height=1.7in]{figures/intergate-sequential.pdf}
%\caption{The two most popular LSTM scheduling schemes for the gate-computation.}
%\label{f:seq_intergate}
%\end{figure}

%\begin{figure}[t!]
%\centering
%\includegraphics[width = 2.8in,height = 1.3in]{figures/unfolded.pdf}
%\caption{Our proposed schedule, \textit{Unfolded} scheme,
%divides the MVMs into the input and hidden parts in order to 
%more efficiently hide across-sequence data dependency.}
%\label{f:unfoldedSchedule}
%\end{figure}

There have been several scheduling approaches proposed in the previous LSTM
implementations. These schemes mainly focus on the different processing order of
the gates~\cite{DBLP:journals/corr/abs-1711-07480, 8416814} and the input and hidden 
vectors~\cite{DBLP:journals/corr/HanKMHLLXLYWYD16, 216077}. 
% in order to overlap the 
% parallel and sequential parts of computation. 
However, they result in sub-optimal resource utilization due
to inefficiently handling of data dependencies because of not pipelining the whole LSTM. 
Existing works mainly focus on speeding up MVMs with a high level of parallelism. 
But according to Amdahl's Law, overall
performance is bounded by the serial execution of the cell-state and hidden units.
To overcome this challenge, we propose \textit{Unfolded} schedule, which removes 
all the parallelism inefficiencies 
by strictly overlapping the dependent and independent parts of computation.

There have been two basic schedulers employed in the previous 
proposals: \textit{Sequential}~\cite{8416814, DBLP:journals/corr/JouppiYPPABBBBB17, DBLP:journals/corr/HanKMHLLXLYWYD16} 
and \textit{Intergate}~\cite{DBLP:journals/corr/abs-1711-07480, DBLP:journals/corr/abs-1803-06305}. 
Figures~\ref{f:unfoldedMotivation}.a and~\ref{f:unfoldedMotivation}.c show the two schemes graphically. 
\textit{Sequential scheduling} computes the gates in a sequential manner, one gate after another, whereas 
\textit{Intergate scheduling} runs all gates' multiplication together by sharing 
MAC resources. Although the two techniques have equal latency in processing the gates 
which includes the MVM and gate activation, there is a slight difference between them. 
\textit{Intergate} scheduling can better hide the latency of 
updating the cell-state and computing the hidden units, by 
pipelining them with the gates' computation (output-based tiling). On the
other hand, \textit{Sequential} scheduling has to wait until reaching the last gate (\textit{Output}) for 
continuing with the cell-state update and producing hidden units.
We will show that the latter schedule outperforms the former in cases that MVM is highly parallelized and 
operates too fast. However, the challenge is that it makes the serial portion of LSTM computation the main bottleneck.

Figure~\ref{f:unfoldedMotivation}.d depicts the \textit{Unfolded} technique graphically.
As illustrated, we first process the input MVM of each time step and save its result in
an intermediate buffer. In other words, we unfold the MVM of the input and hidden vectors in order to hide 
the serialization delay of the recurrent step $t$ with the input MVM of step $t+1$ 
(across-sequence dependency), as there is no dependency between the 
input sequence vectors. 
Then, after accumulating the hidden MVM's output with the buffered input results, 
we can apply the activation function, update the cell-state and generate the hidden outputs.
By processing all gates simultaneously, we can overlap the computation 
of cell-state and hidden units (intra-sequence dependency), by pipelining them in the output-based tiling manner.
By using such computation order, we completely overlap the critical-path delay for evaluating LSTM,
significantly increasing the utilization rate of parallel MAC resources by hiding the two
types of LSTM dependencies shown in Figure~\ref{f:lstmComp}.

Figure~\ref{f:unfoldedMotivation} shows the LSTM critical-path time-line, including the gates' MVMs plus the recurrent serial
computation, for the different scheduling methods. %We assume that the number of resources is equal for all the four approaches.
In order to go over the gate's MVM, we divide the weight matrix into several blocks (shown as red boxes)
to dispatch to the MVM tile-engine. Critical-path is considered as the longest part of the LSTM computation 
between recurrent steps. Due to the dependency of the hidden vector, we cannot completely dispatch 
the next step's MVM while we are processing last sequential portion of the current step.
\textit{Sequential} scheme pipelines each gate's 
computation, including MVM and activation function. 
Since all the four gates' outputs must be ready 
before issuing the cell and hidden update, they run serial to the last gate's MVM. 
\textit{Batch} scheduling is a variant of the previous one, with the difference that it only
processes a batch of each gate at a time, allowing to pipeline the whole LSTM computation.
By issuing all the gates' MVMs at the same time for the \textit{Intergate} approach, we can 
better overlap the computation and decrease the latency for the cell and hidden update by four times. 
Our proposal, \textit{Unfolded} scheduling, not only leverages the advantage of \textit{Intergate} technique 
for hiding the intra-sequence dependency, but it also handles the across-sequence dependency. As
shown in Figure~\ref{f:unfoldedMotivation}.d, while processing the last sequential computation of the cell-state and 
hidden outputs for the current time step, the MAC units can still be busy with calculating 
the next step's input MVM.

We evaluate \textit{Unfolded} schedule on GPU and achieve around 20\% performance improvement compared to the \textit{Sequential} scheme. 
The performance speedup is measured based on the hand-tuned optimized kernels.  
We observe two sources of inefficiencies for GPU architecture, preventing us from 
completely exploiting the \textit{Unfolded} scheduling.
First is that the GPU cores require to pay the cost of synchronization to reduce the partial summation across threads, whereas
this synchronization is implicit in the SHARP’s architecture by using the tree-based structure. 
Second, to parallelize the input/hidden GeMM with the sequential part of LSTM computation, we use two streams to launch these kernels. 
Although, the kernels running on different streams require different hardware resources to do either GeMM (by using Tensor-Cores) or 
the LSTM-Cell update (through CUDA cores), We find out that the GPU hardware resources cannot be efficiently utilized by the two streams. 
On the other hand, this approach is more straightforward to the specific SHARP's acceleration design and through the SHARP's pipeline, we can 
tightly tailor the hardware resources to the type of computation. In order to obtain the performance benefits for the \textit{Unfolded} scheme,
we define a new memory layout, which divides the weight matrix into two partitions of input and hidden. Moreover, as we process all the gates 
together, we put their weights into consecutive parts of the memory based on the tiling dimension selected for MVM processing. 
We go over the different configurations of the MVM tile-engine in the following section.

A similar approach, introduced by~\cite{DBLP:journals/corr/abs-1711-07480,216077}, also tries to partition 
the input and hidden evaluations of LSTM. Their scheme separates the whole input MVMs across all 
the sequence time-steps, focusing on either improving data locality for accessing weight 
matrices~\cite{DBLP:journals/corr/abs-1711-07480} or optimizing LSTM execution through faster
scheduling of GEMMs~\cite{216077}. In contrast, \textit{Unfolded} scheduling unfolds the 
work of each time step individually, and by doing so, we overlap the data-dependency between recurrent
serial processing. Therefore, our mechanism introduces a more efficient pipelining for LSTM in order
to maximize resource utilization.

%Figure~\ref{f:unfoldedMotivation} depicts the four different schedules evaluated in this paper.
%\textit{Sequential} scheme pipelines each gate's 
%computation, including MVM and activation function. 
%Since all the four gates' outputs must be ready 
%before issuing the cell and hidden update, they run serial to the last gate's MVM. 
%\textit{Batch} scheduling is a variant of the previous one, with the difference that it only
%processes a batch of each gate at a time, allowing to pipeline the whole LSTM computation.
%By issuing all the gates' MVMs at the same time for the \textit{Intergate} approach, we can 
%better overlap the computation and decrease the latency for the cell and hidden update by four times. 
%Our proposal, \textit{Unfolded} scheduling, not only leverages the advantage of \textit{Intergate} technique 
%for hiding the intra-sequence dependency, but it also handles the across-sequence dependency. As
%shown in Figure~\ref{f:unfoldedMotivation}.d, while processing the last sequential computation of the cell-state and 
%hidden outputs for the current time step, the MAC units can still be busy with calculating 
%the next step's input MVM.
%\begin{figure}[t!]
%\centering
%\includegraphics[width = 2.5in,height = 1.2in]{figures/unfolded.pdf}
%\vskip -0.5em
%\caption{Our proposed schedule, \textit{Unfolded} scheme,
%divides the MVMs into the input and hidden parts in order to 
%efficiently hide LSTM's data dependency.}
%\vskip -1.5em
%\label{f:unfoldedSchedule}
%\vskip -0.3em
%\end{figure}

%% file: reconfigurability.tex
\section{Improve Adaptability via Reconfigurability}\label{s:lstm_reconnfigurability}

An efficient RNN acceleration design should be able to adapt and scale performance across the space of applications (with different model characteristics) and resource budgets. This means two things: (i) for a fixed resource budget, achieve high-performance execution for different applications (with different model characteristics), and (ii) for a fixed application, scale performance proportionally with resource budget. 

%However, we have not seen this concept of design efficiency achieved in prior work~\cite{8416814, DBLP:journals/corr/JouppiYPPABBBBB17, DBLP:journals/corr/abs-1711-07480}. For example, Figure~\ref{f:brainwave_util} shows that for a fixed resource budget, the BrainWave design~\cite{8416814} achieves high resource utilization for large models, but not for small models. On the other hand, some designs obtain good performance efficiency for a specific model only when resources are limited, but are inefficient with larger resource budgets. For instance, we thoroughly evaluate E-PUR~\cite{DBLP:journals/corr/abs-1711-07480}, 
%the state-of-the-art dense LSTM accelerator, by experimenting across different number of multiply-adder units. Figure~\ref{f:scalability} shows
%the performance improvement obtained by increasing resources when accelerating
%EESEN~\cite{DBLP:journals/corr/MiaoGM15}, one of the benchmarks used 
%in~\cite{DBLP:journals/corr/abs-1711-07480}. As seen, by raising the multiply-add units above 4K, we are not able to achieve an efficient speedup compared to the increase in the number of resources.

In this section, we evaluate SHARP under several configurations for different model characteristics. 
Then, we will show how the design's parameters impact the performance of the system regarding various 
RNN topology. Finally, to improve the adaptability and scalability of our system, we define 
some level of reconfigurability in order to tailor the tile-engine of SHARP's architecture 
to each model.

\subsection{Adaptability Issue}

\begin{figure*}[h]
\centering
\includegraphics[width=\columnwidth]{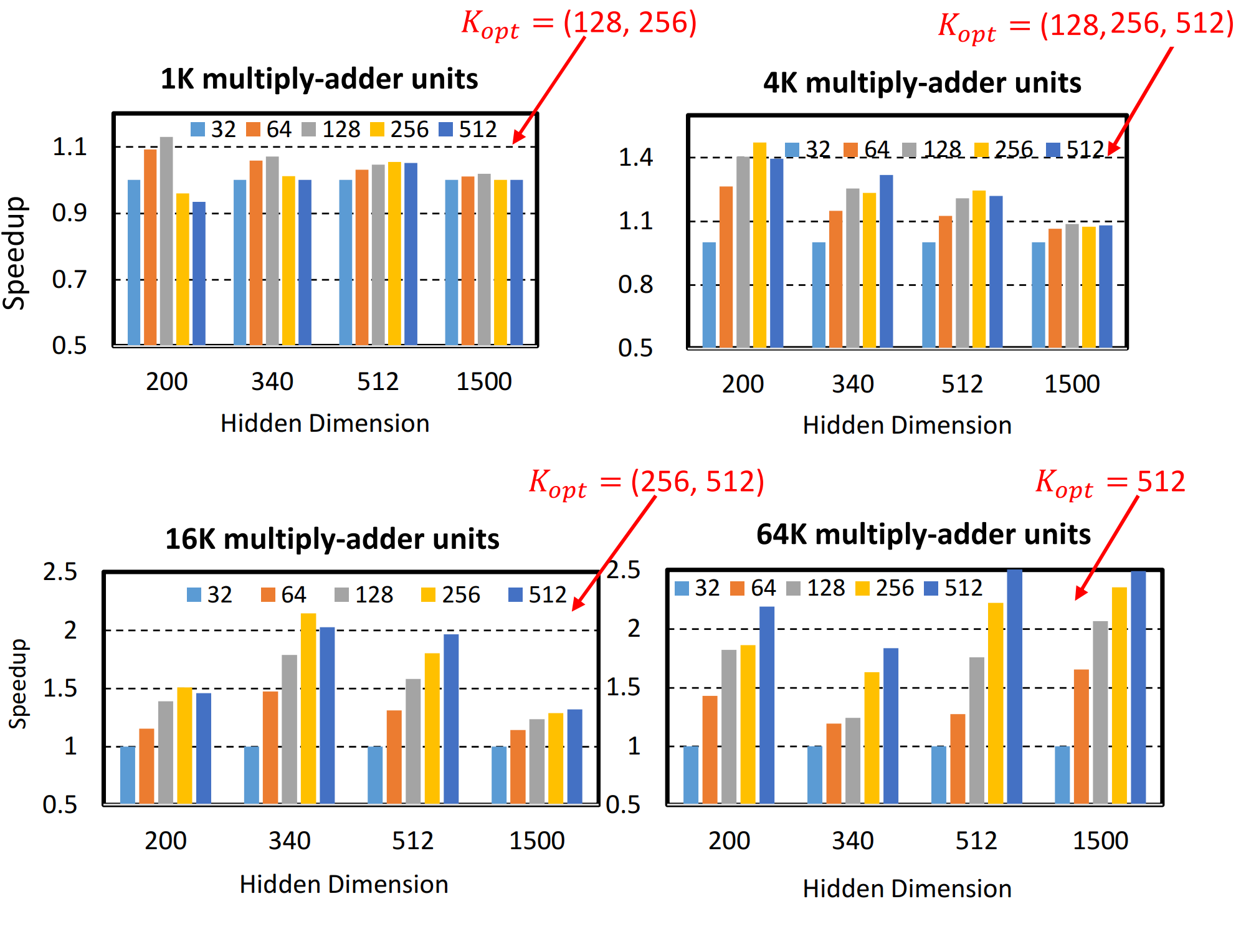}
\vskip -1em

\caption{Exploration on the $K$-width for the VS units of \textit{Compute Unit}. The speedups are normalized to 1K-MAC design. 
In most cases, for different resource budgets, there is not just one configuration providing the best performance for the 
various LSTM dimensions. }
\label{f:Kopt}

\end{figure*}

%\begin{figure}[t!]
%\centering
%\includegraphics[width=2.7in, height=1.2in]{figures/padding_effect.pdf}
%\caption{Padding Effect for the two MVM dispatching techniques: 
%(a) without and (b) with reconfiguration.}
%\vskip -1em
%\label{f:paddingEffect}
%\end{figure}

As explained in Section~\ref{s:pipeline_efficiency}, MVM operations are the main part of RNN pipeline. 
Thus, the way they are assigned to the MAC units defines the pipeline efficiency. However, as we observed,
there is not just one fixed configuration to dispatch weight matrices to the MVM tile-engine. 
This is due to two reasons: first, there is always some padding when tiling the matrix MVM; 
second, there is high performance difference when choosing various tile dimensions for 
an RNN model. Here, we elaborate more on these adaptability issues and then propose reconfiguration at
the MVM tile's architecture, in order to flexibly adapt to each model's requirements and achieve
the highest performance and utilization. 

\subsubsection{MVM Padding}

By using one multiplication tile-engine to go through the whole MVM of the weight matrix, we incur several 
padding due to not fitting the last portion of rows and columns of the matrix into a tile. Therefore,
this results in some resource under-utilization because of not occupying the multipliers that fall out of 
the matrix dimension. 
%Figure~\ref{f:paddingEffect}.a shows the effect of padding using a
%single MVM tile. As depicted in this example, more than half of the second tile-engine does not receive the real
%work to process. 
Furthermore, this padding will continue to exist until multiplying the last column of the matrix.
Note that the only case that padding does not exist is when the size of matrix is a multiple of the tile 
dimension. However, practically speaking, we cannot have as many tile-engines as the different RNN models. 
In order to handle such inefficiency, we apply reconfiguration in 
a way that flexibly changes the tile dimension when reaching the last row segment. 
%Figure~\ref{f:paddingEffect}.b illustrates how reconfiguration can help to remove most
%of the negative effects of the padding, by dynamically changing the tile configuration 
%to dispatch a different number of rows and columns. Therefore, by having the reconfigurability, 
%we utilize the resources more effectively.

\subsubsection{Model Diversity}

As specified in Section~\ref{s:lstm_mvm}, \textit{Compute Unit} is constructed based on
a tile with $K$-row and $N$-column multipliers. Considering the same resources, by choosing different $K$ widths, 
the tile dimension (rows and columns) varies, therefore results in various latency to complete the $K$ partial results of MVM. 
For instance, if $K$ is too small (\textit{Config4} at Figure~\ref{f:mvmConfigs}), we place the multipliers more column-wise, 
producing partial results faster than the case that $k$ is too large(\textit{Config1} at Figure~\ref{f:mvmConfigs}). 
%We have explored several widths of the VS units for performing different 
%LSTM hidden/input dimensions. Moreover, we consider a range of resource budgets from 1K to 64K MACs. 
Figure~\ref{f:Kopt} shows the $k$-width exploration results in four charts corresponding to 
1K, 4K, 16K and 64K multiply-adders, respectively. Each chart illustrates the performance 
evaluation of choosing several $K$ widths from 32 to 512, regarding
the different LSTM hidden dimensions. Note that we assume equal size for both the hidden and input vectors in
our experiments. Moreover, we run all the models for the same sequence length of 25 time steps. As seen, 
there is not just one best configuration for tiling the MVM operations. For instance, in the
case of 4K MAC units, there are different optimal $K$-widths (128, 256, and 512) for each LSTM dimension 
that result in the highest performance speedup. 

%By selecting $k$ , we configure the latency at each LSTM pipeline stage as we choose the 
%number of results produced at each gate. This directly impacts the pipeline efficiency for overlapping different computation parts. Furthermore,
%in order to have a fast dispatching to the parallel multipliers, SHARP satisfies the
%data-dependencies by hiding the recurrent serialization of LSTM processing.
%As various models have different ratio between the computation of each time step, MVMs and 
%the recurrent connections (cell-state update and hidden generation), they require different resource mapping 
%in order to balance the latency between the pipeline stages. As a result, we obtain a flexible architecture by
%adapting its configuration for each LSTM model.

\subsection{Re-configurable Compute Unit}

In order to increase the adaptability of our design, we modify the \textit{Compute Unit}
component (as shown in Figure~\ref{f:recTreeAdder}) in a way to configure the MVM tile-engine 
based on each LSTM model dimension. Initially, we consider 32 as the $K$-width of each VS unit. 
Then, by mapping the VS units at either rows or columns of the weight matrix, we can generate different 
MVM tiles. Figure~\ref{f:mvmConfigs} depicts the four possible tile configurations of SHARP. %For instance, if
%we want to process 256 rows at each tile, we use 8 basic VS units at the row dimension,
%while placing the rest ($N/8$) at the column direction. 
Even though the number of multiplications does not differ at each configuration, the data 
dispatching pattern should match to the row and column selection scheme. This also affects the
number of partial results generated by the \textit{R-Add-Reduce} stage. Therefore, we rearrange the 
memory organization of the weight matrix by interleaving them based on the configured tile dimension. 
We also interleave the weights in a way to keep all the VS units uniformly busy.

After delivering the multiplication results to the \textit{R-Add-Reduce} stage, the tree-adder uses
a reconfigurable routing mechanism in order to emit the correct partial results corresponding to
the MVM tile configuration. By employing four multiplexers, we support the four configurations 
shown in Figure~\ref{f:mvmConfigs}, by selecting between the outputs of the last four levels of the tree-adder. 
Figure~\ref{f:recTreeAdder} illustrates the configurations in different colors and also the 4 multiplexers 
we use to reroute adders' outputs to match the tiling topology. For example, in the case of \textit{Config1} of 
Figure~\ref{f:mvmConfigs}, we select eight partial sums from the fourth last level ($LogN-3$) of the tree-adder to 
send to the accumulators. Then, by reaching to the end of input and hidden vectors, we will have $8\times K$ MVM results.
SHARP's controller multiplexes based on each model's specification, by configuring the bit-selects form a  
table that stores them for the different LSTM dimensions.

\subsubsection{Impact on Padding}

\begin{figure}[t!]
\centering
\includegraphics[width=3.345in]{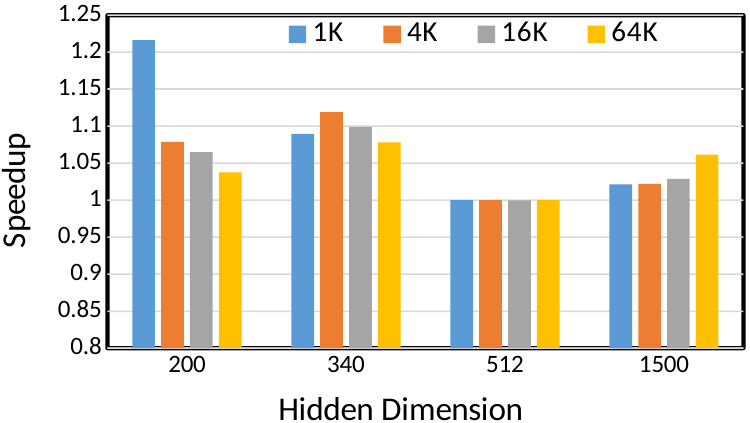}
\vskip -0.5em
\caption{Performance speedup achieved using the padding reconfiguration technique
regarding different resource budgets and for the various LSTM dimensions.}
\label{f:paddingReconfig}
\vskip -1.2em
\end{figure}

In order to measure the effectiveness of our scheme, we have evaluated padding reconfiguration 
for the different LSTM models and similar range of resources as for Figure~\ref{f:Kopt}. 
Regarding the MVM tile-engine, we configure $K\_opt$ for each combination of 
LSTM dimension and MAC resources. Then, we compare the accelerator's performance
for the two cases, applying fixed or reconfigurable configurations. Note that
the controller reconfigures the tile-engine dynamically, in a way that $K$ gets as close as to the remaining 
number of rows. Figure~\ref{f:paddingReconfig} shows the speedups achieved for SHARP by using
reconfigurability when running various LSTM models, considering different MAC units. 
As seen, we improve performance in almost all the cases, except for 512 hidden dimension.
The reason is that as 512 is a multiple of $K\_opt$, it causes no padding during
MVM tiling, and hence, there is no benefit of reconfiguration. In total, 
we can get up to 1.22x speedup by applying our approach to alleviate padding.

\subsubsection{Impact on Adaptability}

By using the reconfigurability, we can generate almost all the $K\_opt$ widths for the
MVM tile-engine regarding the LSTM models shown in Figure~\ref{f:Kopt}, by combining the 
basic 32-width VS units. We can select between the four options from 32 to 256 for the $K$, achieving 
most of the performance benefits considering a variety of model dimensions. Note that we 
explore the configurations offline in order to determine the parameters that reach the best performance for
each application. 
This generates a table with several entries, each storing the optimal configuration for 
each LSTM's hidden dimension, including the multiplexing and control-logic applied for the tree-adder.
The table is preloaded in an on-chip memory in SHARP, minimizing the cost of reconfiguration 
both performance- and energy-wise.

Reconfiguration in SHARP has negligible runtime cost and it works as follows. Prior to 
the execution of each LSTM layer, its optimized architecture's configuration is fetched from the 
aforementioned table. Then, SHARP sets the control signals of the multiplexers for the tree-adder 
accordingly, which has negligible performance overhead. 
Note that the expensive operations such as experimentally finding the optimal 
configuration or changing the memory layout are performed offline. Runtime operations only include an access 
to a small table and setting the control signals of several multiplexers. 

For every LSTM network, its weights are rearranged and interleaved offline according to the access-pattern 
of the optimal configuration. Then, we fetch the memory blocks likewise in order to fill in the on-chip buffers,
the SRAM banks corresponding to the VS units. Except for the initial delay to fetch the memory requests 
(this delay is proportional to the model's size and the LSTM dimension), we can overlap the rest with the 
computation of MVM tile-engine. Furthermore, after having the weight matrices 
reside on-chip, there will be no off-chip memory bottleneck restricting SHARP's performance. 
Regarding the input sequences, the I/H buffer (see Figure~\ref{s:lstmDesign}) works in a ping-pong 
manner. While the MVM tile-engines are processing the current batch of data, SHARP prefetches 
the next part of input data.

In order to measure the effectiveness of reconfigurability, we compare the configured $K$ against an 
ideal case of using a hardened $K$ for each model. Our numbers show very similar performance
evaluation as all the multiplexing latency is covered by the \textit{R-Add-Reduce} slack time, 
imposing no extra cycle to send out the partial sums. The only overhead is for the reconfiguration logic
at the controller to select the best configuration, which only happens at the beginning of each LSTM
layer's computation. 

By combining the reconfigurability with our scheduling technique, we improve the scalability
of our architecture by increasing the performance efficiency with high utilization of the available 
resources. More specifically, by carefully pipelining the LSTM computation and balancing 
the latency of pipeline stages, we can overlap most of the data dependencies that limit the 
parallelism of the MVM operations. Furthermore, comparing with the previous methods that 
show a poor scaling factor by increasing the number of resources (see Figure~\ref{f:scalability}), 
we significantly improve the utilization by handling the serialization of LSTM computation more efficiently. 
For instance, as seen in Figure~\ref{f:Kopt}, the speedup numbers are relatively higher in the 
cases of using more resource budgets such as 16K/64K.

%% file: methodology.tex
\begin{table}[t!]

\caption{SHARP's configuration.}

\label{t:acc_hw_params}

\centering
\begin{tabular}{|c|c|}
\hline
\cellcolor[gray]{0.9}\small Technology - Frequency&\small 32 nm - 500 MHz\\
\cellcolor[gray]{0.9}\small Weight - I/H Buffer&\small 26 MB - 2.3 MB\\
\cellcolor[gray]{0.9}\small Cell State&\small 192 KB\\
\cellcolor[gray]{0.9}\small Intermediate Buffer&\small 24 KB\\
\cellcolor[gray]{0.9}\small MFUs&\small 64 Units\\
\cellcolor[gray]{0.9}\small Multiplication precision&\small 16-bit Floating-Point\\
\cellcolor[gray]{0.9}\small Multiply-Adder Units&\small 1K , 4K , 16K , 64K\\
\cellcolor[gray]{0.9}\small Peak Bandwidth (GB/s)&\small 11 , 44 , 170 , 561\\
\cellcolor[gray]{0.9}\small Peak Throughput(TFLOPs) &\small 0.46 , 1.86 , 7.4 , 29.8\\ 
\hline
\end{tabular}
\end{table}

\section{Evaluation Methodology}\label{s:EvaluationMethodology}

In order to evaluate SHARP, we developed an architectural C++ cycle-accurate simulator to
accurately model all the pipeline stages described in Section~\ref{s:lstmDesign}. SHARP's design 
is configured using the parameters shown in Table~\ref{t:acc_hw_params}. Because we consider different 
resource budgets from 1K to 64K MACs, we can obtain a range of peak throughput between 0.46 and 29.8 TFLOPS/s. 
However, by increasing the resources, we require higher peak bandwidth from the on-chip memory components, 
up to 561 GB/s for the 64K-MAC configuration. To achieve this bandwidth, we increase the 
banks of SRAM buffers proportional to the VS units of SHARP's architecture.

In order to estimate the latency, power and area of our design, we implemented all the logic components 
in Verilog using the Synopsys Design-Ware library~\cite{s111}. Then, we synthesized them with 
Synopsys Design Compiler using the commercial 32 nm technology library~\cite{s222}.
For the technology library, we used 0.85 V power configuration and the typical-typical process corner.
Furthermore, We modeled the on-chip SRAM buffers in CACTI-P~\cite{Cacti} with the same technology parameters. 
The simulator incorporates the timing of the critical-path-delay from the synthesis results and memory latencies.
Finally, to model the off-chip DRAM main memory, we use the Micron Power model for an 8-GB LPDDR~\cite{TN-41-01}.
The cycle-accurate simulator provides the activity factor for the diﬀerent components and the total cycle count, which
are used to compute execution time, dynamic and static energy by combining them with the estimations of the Design Compiler,
CACTI and Micron power models.
%The simulator counts the number of cycles for executing 
%LSTM computations of an input sequence, while accounting for each component's delays.

Table~\ref{t:area_res} shows the area breakdown for the different versions of SHARP.
For the 1K-MAC configuration, more than 86\% of the area is allocated by the SRAM buffers. 
However, as we increase the number of MACs, the compute-unit becomes the dominant part of the
accelerator's area. This is because we use half-precision for the multiplication and full-precision
for the accumulation. The area usage can be significantly reduced by using different low-precision
schemes such as K-Means~\cite{10.2307/2346830} or linear quantization, as shown in the previous works~\cite{htabani, songi, masr}. By adding the reconfigurability,
we only add less than 2\% overhead in the Add-reduce module and lower than 0.1\% in the total area of accelerator.

To set the frequency of the system, we consider the critical path delay and access times 
reported by Design Compiler and CACTI, respectively. We take the maximum delay among the different 
components, which is 1.94 ns for the half-precision (16-bit) multiplication, resulting in nearly 
500 MHz frequency.

\begin{table}[t!]

\caption{Area breakdown of different configurations of SHARP.}

\label{t:area_res}

\centering
\begin{tabular}{|c|c|c|c|c|}
\hline
\cellcolor[gray]{0.9} \small \textbf{Number of MACs} & \small \textbf{1K}& \small \textbf{4K}& \small \textbf{16K}& \small \textbf{64K}\\
\hline
\cellcolor[gray]{0.9} \small \textbf{Compute-Unit (\%)}&\small 7.4&\small 22.4&\small 52.6&\small 80.9\\
\hline
\cellcolor[gray]{0.9}\small \textbf{SRAM Buffers (\%)} & \small 86.2& \small 72.7& \small 44.3& \small 17.6\\
\hline
\cellcolor[gray]{0.9}\small \textbf{MFU + Cell-Updater (\%)} & \small 6.3& \small 4.7& \small 2.8& \small 1.08\\
\hline
\cellcolor[gray]{0.9}\small \textbf{Pipeline Controller (\%)} & \small 0.09& \small 0.1& \small 0.3& \small 0.4\\
\hline
\cellcolor[gray]{0.9}\small \textbf{Reconfiguration Logic (\%)} & \small 0.08& \small 0.06& \small 0.04& \small 0.02\\
\hline
\cellcolor[gray]{0.9}\small \textbf{Total Area ($mm^2$)} & \small 101.1& \small 133.3& \small 227.6& \small 591.9\\
\hline
\end{tabular}
\end{table}

Regarding the previous implementations, we implemented E-PUR scheduling~\mbox{\cite{DBLP:journals/corr/abs-1711-07480}} 
by modifying SHARP's architecture in order to enable a thorough comparison of our design with the state-of-the-art 
ASIC-based LSTM acceleration. Moreover, since BrainWave is not open sourced, we developed a cycle-accurate performance model for the BrainWave
FPGA implementation~\mbox{\cite{8416814}}. Similar to previous designs~\cite{8416814,DBLP:journals/corr/abs-1711-07480}, 
we assume the input-features and model-parameters already 
exist in the main-memory before the accelerator begins the LSTM processing.
% that counts the number of cycles for the LSTM computation, accounting 
%for all the pipeline stalls and memory accesses. 
We validated the correctness of our model, by comparing against the number of cycles reported in~\cite{8416814},
using the Structurally-Constrained Model Critical-Path analysis.
In order to have a fair comparison, our BrainWave implementation does not account for the network latency. 
%Therefore, 
% with our accelerator 
%our reproduced numbers are better than what was 
%reported in~\mbox{\cite{8416814}}.
% because we are not accounting for the network latency in order to have a fair 
% comparison with our accelerator.

The LSTM hidden dimensions are selected from the LSTM networks of popular applications such as machine reading
comprehension~\mbox{\cite{DBLP:journals/corr/SeoKFH16}}, language modeling~\mbox{\cite{DBLP:journals/corr/ZarembaSV14}}, 
speech recognition~\mbox{\cite{DBLP:journals/corr/MiaoGM15}}, and machine translation~\cite{DBLP:journals/corr/WuSCLNMKCGMKSJL16}.

%% file: results.tex
\section{Experimental Results}\label{s:ExperimentalResults}

This section presents an experimental evaluation of SHARP, by measuring the impact of our 
proposed techniques on improving performance in terms of: (1) reducing execution time and increasing 
resource utilization, and (2) reducing energy consumption. Here, we present the result of LSTM as the most
well-known RNN model, however, the same improvement can be achieved in other networks that have similar design
, such as GRU.
First, we show the latency and utilization of the different LSTM's configuration for several hidden dimensions. Next, we compare our numbers 
with different state-of-the-art systems. Then, we show the energy consumption of SHARP for various 
scheduling approaches. Finally, we report some results on the power breakdown.

\begin{figure}[t!]
\centering
\includegraphics[width=4in]{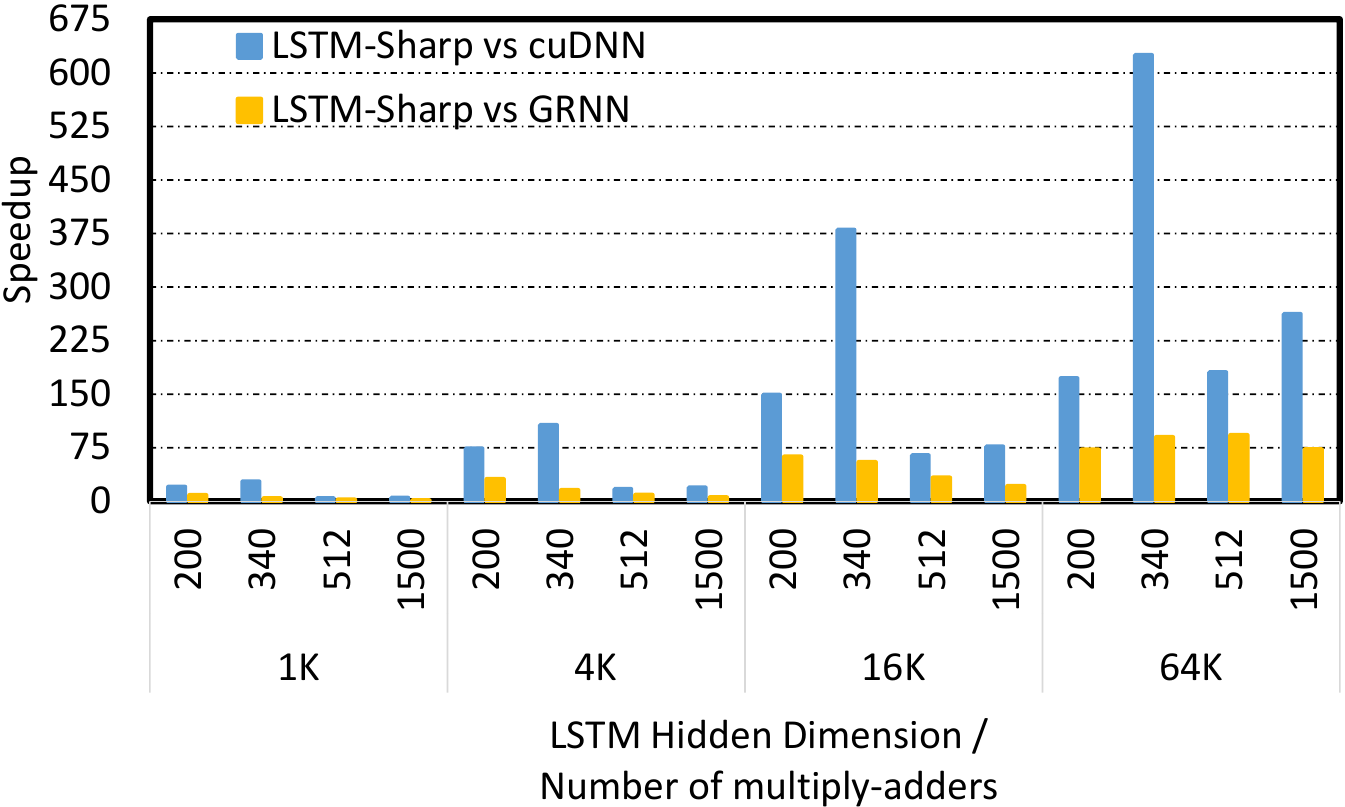}
\caption{Performance evaluation of the four schedulers for the different LSTM 
models and resource budgets. We consider sequence-length as 25 in all cases and 
similar size for hidden and input vectors. Speedup
numbers are normalized to \textit{Sequential} scheduling. }
\label{f:scheduling_cmp1}
\end{figure}

Figure~\ref{f:scheduling_cmp1} shows the performance comparison of the schedulers discussed in Section~\ref{s:lstm_analysis}. 
Each set of 4 bars shows the speedups normalized to the first bar (\textit{Sequential} scheduling).  
We evaluate the numbers considering all resource budgets and the LSTM models used for experimenting
the design. Regarding the MVM tile-engine ($N\times$ $k$-width VS units), we consider 
32 rows for the $k$ and mapping all the VS units to the columns of the weight matrix.

As Figure~\ref{f:scheduling_cmp1} depicts, \textit{Unfolded} scheme always obtains the best performance of all, 
since it removes most of the data-dependencies and highly utilizes the parallel MACs. However, the benefit diminishes
by increasing the LSTM dimension or reducing the number of MACs. The reason is that MVMs
become the main performance bottleneck under those conditions, and hence, the way we order
the LSTM computation cannot have the expected impact. On the other hand, \textit{Intergate} scheduling 
outperforms \textit{Sequential} and \textit{Batch} scheduling.
Comparing with our proposal, it provides less benefit as it only removes the intra-sequence
dependency whereas across-sequence dependency still remains. \textit{Batch} and \textit{Sequential} schedules 
show almost similar execution, due to not efficiently handling all the data-dependencies.

\begin{figure}[t!]
\centering
\includegraphics[width=5in]{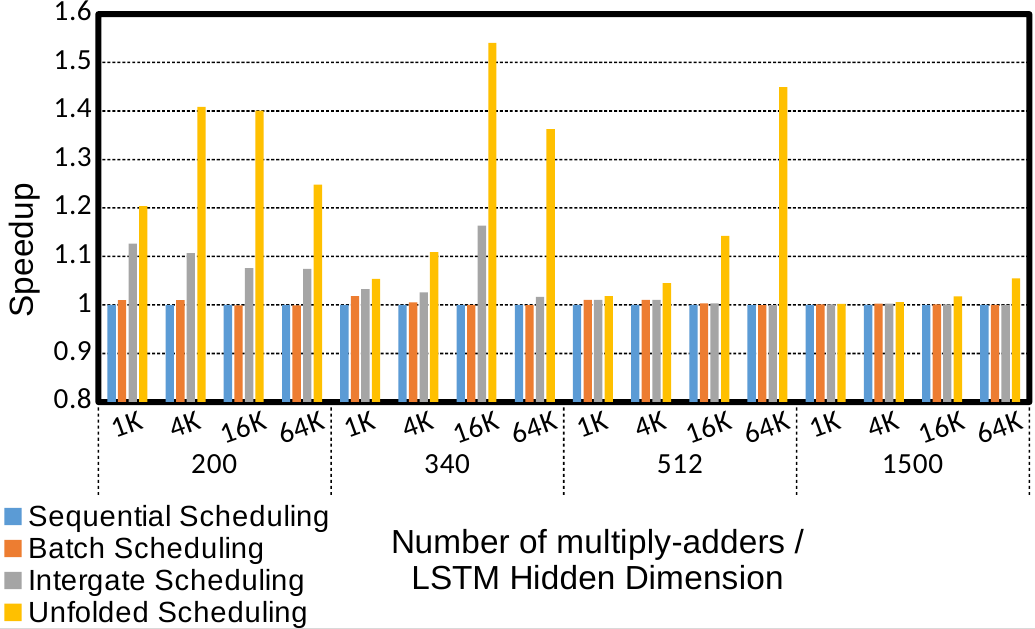}
\caption{Latency and resource-utilization of SHARP for different LSTM 
models and resource budgets. We consider sequence-length as 25 in all cases and 
similar size for hidden and input vectors. }
\label{f:scheduling_cmp}

\end{figure}

Figure~\ref{f:scheduling_cmp} shows the execution time and resource-utilization considering the 
different resource budgets for SHARP and the LSTM models used for the design experimentation.
The MVM tile-engine ($N\times$ $k$-width VS units) is configured based on the exploration result 
shown in Figure~\ref{f:Kopt}. Moreover, the tile is dynamically reconfigured
to reduce the padding impact. As depicted, SHARP scales well as it linearly reduces the execution time (AVG case)
by increasing the number of MACs from 1K to 64K. Furthermore, we obtain relatively high utilization for
all the cases, ranging from 50\% to 98\% for 64K- to 1K-MAC resource-budgets, respectively.
Compared to the ASIC-based EPUR architecture, resulting 24\%, 49\%, 74\%, and 95\% utilization 
for 1K-64K configurations, our design scales better for a variety of resource-budgets, 
particularly when increasing MAC units (1.3x - 2x).
These performance benefits come form both the efficient scheduling as well as the 
reconfigurability of SHARPs' architecture. With scheduling, we efficiently distribute the 
workload among different pipeline stages, whereas by reconfigurable MVM-tiling, we decide on
the amount of work assigned for each stage. Consequently, by relaxing the two main dependencies of LSTM-computation, 
we are able to manage the resources more effectively.

\begin{figure}[t!]
\centering
\includegraphics[width=4in]{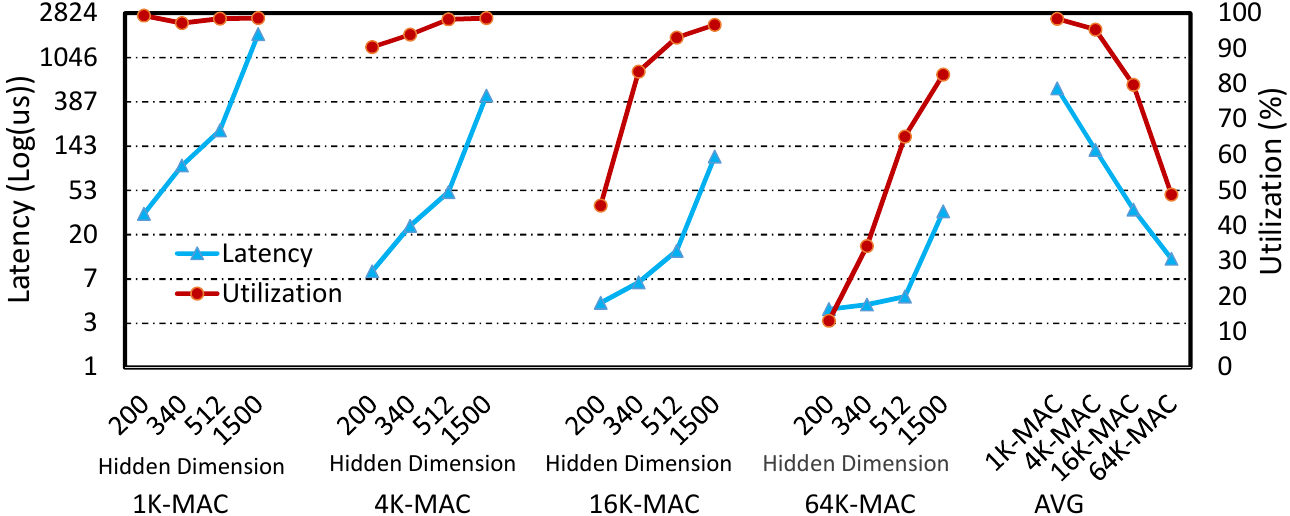}
\caption{SHARP's speedup versus the most recent GPU implementations. }
\label{f:speedupVsGPU}
\end{figure}

We compare SHARP against the state-of-the-art GPU, FPGA and ASIC implementations, i.e. cuDNN~\cite{DBLP:journals/corr/ChetlurWVCTCS14}, 
GRNN~\cite{Holmes:2019:GLS:3302424.3303949}, BrainWave~\cite{8416814} and 
E-PUR~\cite{DBLP:journals/corr/abs-1711-07480}. We show the hardware specification of each of these platform in Table ~\ref{t:hardwarespec}.
Figure~\ref{f:speedupVsGPU} shows the speedup achieved for the SHARP compared to the most recent scalable, efficient implementations on GPU. We use NVIDIA Titan V for the GPU evaluation that has a theoretical peak throughput of 29.8 TFLOPs (FP16). 
Speedup numbers are reported for all the different configuration,
executing the various LSTM models that we have tested. As seen, our ASIC implementation outperforms 
the GPU's in almost all the cases by up to 1 to 2 orders of magnitude. 
Considering 64K-MAC configuration, which has equal peak throughput as Titan V, SHARP
obtains 172-625x and 72-93x faster LSTM inference than the cuDNN and GRNN GPU implementations.

\begin{table}[t!]
\caption{Hardware Configurations.}
\vskip -0.5em
\label{t:hardwarespec}
\centering
\begin{tabular}{|c|c|c|c|c|}
\hline
\cellcolor[gray]{0.9}\small \textbf{Architecture}&\small \textbf{\#cores}&\small \textbf{Clock Speed(MHz)} &\textbf{Compute Precision} & \textbf{Power (W)} \\
\hline
\cellcolor[gray]{0.9} Titan V& 5120& 1200& float16& 250\\
 %33150/6152
\hline
\cellcolor[gray]{0.9} BrainWave& 96000& 250& float16& 125\\
\hline
\cellcolor[gray]{0.9} EPUR& 1024& 500& float16& 1\\
 %5525/4483
\hline
\end{tabular}
\end{table}

Table~\ref{t:brainwave} shows the performance speedup achieved by our accelerator compared to the Stratix-10 
version of BrainWave. Note that we reduce SHARP's frequency from 500 MHz to 250 MHZ, similar to the BrainWave's design, so 
as to have a fair comparison. In addition, we increase our MACs to 96K to have equal resource budget. 
We choose the LSTM configurations form Baidu DeepBench~\cite{Baidu_DeepBench}, similar to BrainWave's experiments.
As seen, we achieve more than 1.65x speedup for all the LSTM models and the speedups are significantly larger for the 
smaller dimensions. This shows that we alleviate the adaptability issue of BrainWave (see Figure~\ref{f:brainwave_util}).

\begin{table}[t!]
\caption{DeepBench LSTM Inference speedup over BrainWave~\cite{8416814}.}
\vskip -0.5em
\label{t:brainwave}
\centering
\begin{tabular}{|c|c|c|}
\hline
\cellcolor[gray]{0.9}\small \textbf{LSTM hidden dimension}&\small \textbf{Time-steps}&\small \textbf{Speedup (X)}\\
\hline
\cellcolor[gray]{0.9} 256& 150& 5.39\\
 %33150/6152
\hline
\cellcolor[gray]{0.9} 512& 25& 3.57\\
\hline
 \cellcolor[gray]{0.9}1024& 25& 1.85\\
 %5525/4483
\hline
\cellcolor[gray]{0.9} 1536& 50& 1.73\\
\hline
\end{tabular}
\end{table}

Furthermore, we compare LSTM performance considering several real-world application in Speech Recogniton, Video Captioning and also Machine Translation.
Table~\ref{t:lstm_config} shows the configuration of the different networks used for measuring the performance evaluation. 
Table~\ref{t:epur} shows the speedup SHARP achieves when running each network compared to E-PUR for the different number of MACs. Note that we use the same clock frequency of 500 MHz for both EPUR and SHARP. 
The results show that we obtain relatively higher speedups as we increase the number of resources. 
As a result, we improve the scalability of RNN acceleration for a range of resource budgets and different models.

\begin{table}[t!]
\caption{LSTM Network's Configuration}
\vskip -0.5em
\label{t:lstm_config}
\vskip -0.5em
\centering
\begin{tabular}{|c|c|c|c|c|}
\hline
\cellcolor[gray]{0.9}\small \textbf{Benchmark} &\small \textbf{Layers}&\small \textbf{LSTM} &\small \textbf{Hidden} & \small  \textbf{Time} \\
\cellcolor[gray]{0.9}\small \textbf{} &\small \textbf{}&\small \textbf{Type} &\small \textbf{Units} & \small  \textbf{Steps} \\

\hline
\cellcolor[gray]{0.9}\small {EESEN~\cite{DBLP:journals/corr/MiaoGM15}}& \small 5 & \small Bi-dir & \small340 & \small300-700 \\
\hline
\cellcolor[gray]{0.9}\small {GMAT~\cite{wu2016googles}}& \small17& \small\small Uni-dir& \small1024& \small50-100\\
\hline
\cellcolor[gray]{0.9}\small {BYSDNE~\cite{ng2015short}}& \small5 & \small Uni-dir & \small340 & \small30\\
\hline
\cellcolor[gray]{0.9}\small {RLDRADSPR~\cite{kim2017residual}}& \small10& \small\small Stacked& \small1024& \small300-512\\
\hline
\end{tabular}
\end{table}
\begin{table}[t!]
\vskip -0.5em
\caption{SHARP's speedups w.r.t. EPUR~\cite{DBLP:journals/corr/abs-1711-07480} }
\vskip -0.5em
\label{t:epur}
\centering
\begin{tabular}{|c|c|c|c|c|}
\hline
\cellcolor[gray]{0.9}\small \textbf{Number of MACs} &\small \textbf{1K}&\small \textbf{4K}&\small \textbf{16K}&\small \textbf{64K}\\
\hline
\cellcolor[gray]{0.9}\small \textbf{EESEN}& 1.07& 1.25& 1.68& 1.9\\
\hline
\cellcolor[gray]{0.9}\small \textbf{GMAT}& 1.01& 1.51& 1.53& 1.66\\
\hline
\cellcolor[gray]{0.9}\small \textbf{BYSDNE}& 1.05& 1.24& 1.8& 2.22\\
\hline
\cellcolor[gray]{0.9}\small \textbf{RLDRADSPR}& 1.03& 1.11& 1.45& 2.3\\
\hline
\end{tabular}
\end{table}

Furthermore, we measure the energy consumption for both SHARP and E-PUR considering different number of resources while using the same clock frequency. 
Figure~\ref{f:energy_cmp} shows the energy consumption, normalized to E-PUR using 1K MACs, for the different 
LSTM dimensions. As it illustrates, SHARP obtains better energy-efficiency for the smaller models when there
is lower resources available (1K- and 4K-MAC). This is due to the effectiveness of both \textit{Unfolded} scheduling 
as well as flexible tiling. Moreover, reconfiguration handles the padding of LSTM's MVM which is critical 
for the performance of smaller models and when there is less number of MACs (Figure~\ref{f:paddingReconfig}). 
Moreover, we achieve higher energy-reduction for larger SHARP's designs, as we achieve better scalability than E-PUR (Table~\ref{t:epur})
through efficiently utilizing SHARP's MAC resources. Note that even though we increase power dissipation by between 1.4\% to 36\%, 
as the LSTM inference takes less time, therefore our energy numbers, which is $power \times time$, decreases more.
%Considering the average energy breakdown, static energy accounts for 46.4\% to 28\%
%regarding the different SHARP's configuration using 1K to 64K MACs, respectively. 
In total, we reduce SHARP's energy consumption 
on average by 7.3\%, 18.2\%, 34.8\%, and 40.5\% when using 1K to 64K MACs, respectively.

Figure~\ref{f:power-breakdown} depicts the breakdown of power dissipation among the main
components of our design. We consider an average case to compute the power for the different
resource-budgets. As the figure illustrates, a large share of the power goes to the SRAM buffers,
which is the dominant consumer for 1K- and 4K-MAC configurations. On the other hand, as we increase
the number of resources, the compute-unit dissipates more power. Furthermore, main memory consumes
more power and energy as the number of MACs grows, due to the high bandwidth requirement.
The activation part takes almost similar amount of power as it does not change between different designs. Finally, the pipeline controller which also includes the reconfiguration logic has less than 1\% share of the total power.

%% file: related_work.tex
\section{Related Work}\label{s:RelatedWork}

\begin{figure}[t!]
\centering
\includegraphics[width=5in]{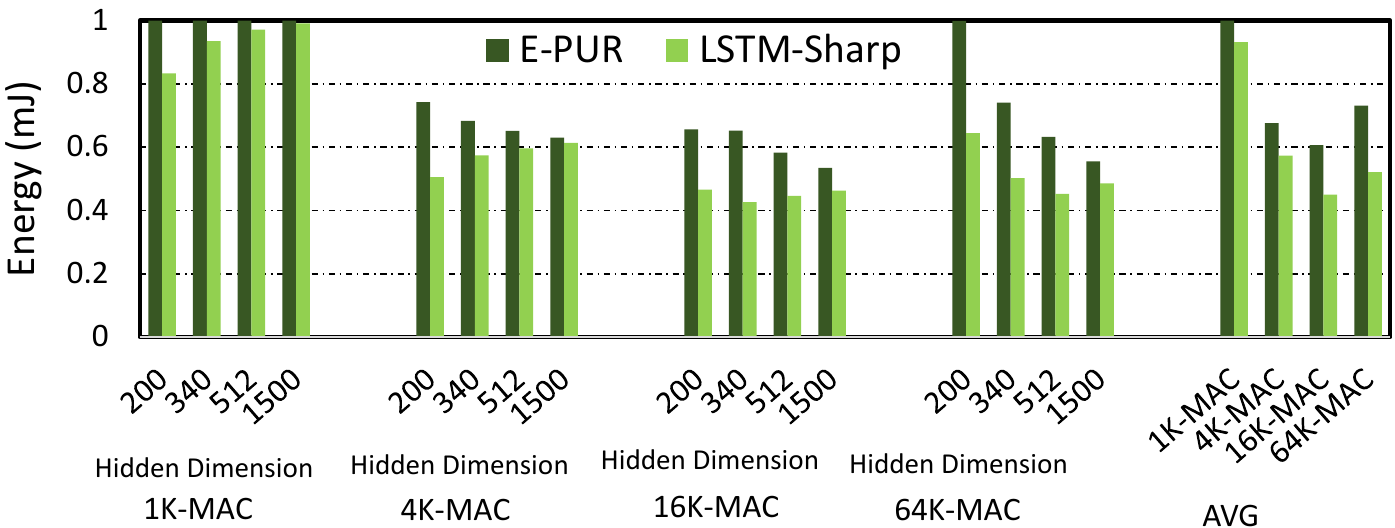}
\caption{Energy consumption of the SHARP for different hidden dimensions. The numbers are
normalized to E-PUR with 1K-MAC resources.}
\vskip -1em
\label{f:energy_cmp}
\end{figure}

\begin{figure}[t!]
\centering
\includegraphics[width=3.345in]{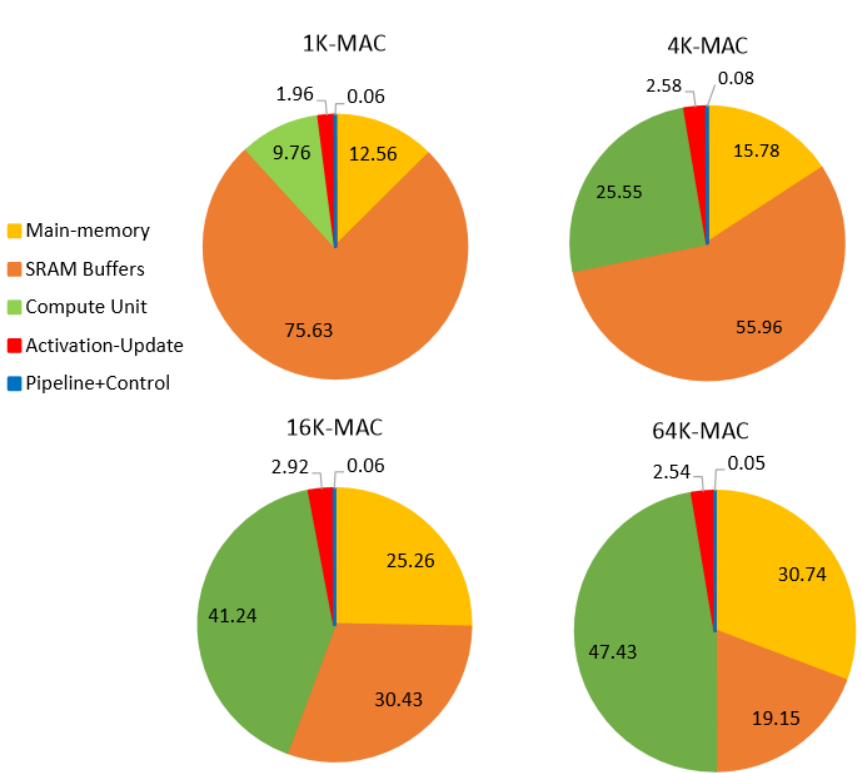}
\caption{Power breakdown of SHARP for different resource
configurations. We average the percentages for running different applications.
SHARP consumes 8.11, 11.36, 22.13, 47.7 Watts of power under 1K to 64K MACs.}
\vskip -0.5em
\label{f:power-breakdown}
\end{figure}
To optimize the performance and energy of RNNs, a plethora of customized 
architecture~\cite{DBLP:journals/corr/abs-1711-07480,DBLP:journals/corr/HanKMHLLXLYWYD16, DBLP:journals/corr/abs-1803-06305} 
and neural-processing~\cite{DBLP:journals/corr/JouppiYPPABBBBB17, 8416814} designs 
have been recently proposed 
in order . 
Most of the previous accelerators' implementation are FPGA-based~\cite{DBLP:journals/corr/ChangMC15, 23083700, 8416814},
whereas only a few works have been explored for ASIC design~\cite{DBLP:journals/corr/JouppiYPPABBBBB17,DBLP:journals/corr/abs-1711-07480}. 
Furthermore, these designs are either targeted for mobiles and wearables~\cite{DBLP:journals/corr/abs-1711-07480}, 
or data centers and cloud networks~\cite{8416814, DBLP:journals/corr/JouppiYPPABBBBB17}. The latter cases
normally restrict themselves to a relatively small resource budget, whereas the former employ a large
amount of parallelism. SHARP aims to optimize RNN for a variety of design points and
considering different model characteristics.

We have gone through some of the previous works~\cite{8416814, DBLP:journals/corr/JouppiYPPABBBBB17, DBLP:journals/corr/abs-1711-07480} 
throughout the paper and compare them against SHARP's evaluation results in Section~\ref{s:ExperimentalResults}.
The rest implementations focus on two main problems for RNN acceleration: optimizing the inference 
computation~\cite{7858394, 7160054}, and reducing memory requirements by compression and pruning techniques~\cite{DBLP:journals/corr/HanKMHLLXLYWYD16,DBLP:journals/corr/abs-1803-06305}. 
%While most of these architectures concentrate on accelerating dense networks, EIE~\cite{han2016eie} and 
%ESE~\cite{DBLP:journals/corr/MiaoGM15} are the two closely 
%related works that optimize LSTM through execution engines for sparse networks.
%C-LSTM~\cite{DBLP:journals/corr/abs-1803-06305} is another 
%alternative working on compressed LSTM models using FFT-based block-circulant matrices.

In this paper, we mainly target dense and uncompressed RNN models. Furthermore,
we consider that all the synaptic weights fit on-chip for one layer execution, similar to
E-PUR~\cite{DBLP:journals/corr/abs-1711-07480} and BrainWave~\cite{8416814}. 
Other designs, such as ESE~\cite{DBLP:journals/corr/MiaoGM15} and C-LSTM~\cite{DBLP:journals/corr/abs-1803-06305}, 
take another approach that tries to pipeline the memory fetches with the LSTM computations in order
to hide the latency of accessing off-chip data. However, we observe that such design schemes
cannot provide good scalability and performance efficiency with high amount of parallelism, resulting in
low utilization. We build SHARP
in a way to optimize for various purposes from wearables to mobile systems and cloud networks.
Also, we ensure that it achieves a scalable performance for a range of models.
%On the other hand, NPUs are mainly designed for performing a range of neural
%networks, and therefore are not customized to the needs of each one. 

Apart from FGPA and ASIC, researchers and practitioners are also looking into using GPUs for 
RNN~\cite{persistent/rnn/baidu, munich-cuda-rnn, cuDnn}. They are mainly designed for maximizing training
throughput, whereas the GPU utilization is limited by insufficient
parallelism when the model size and batch size
are small, which is difficult to make full usage of massive GPU cores.

%% file: conclusion.tex
\section{Conclusions}\label{s:Conclusions}

In this paper, we propose SHARP, a scalable, high-performance RNN accelerator,
which consists of a novel scheduling scheme --- \textit{Unfolded} that strictly overlaps the LSTM computation by hiding almost
all the dependencies, and a dynamically reconfigurable
architecture to improve the adaptability.
SHARP tailors the resource allocation based on the requirements of a particular model. 
It achieves significant performance speedup, energy-saving and utilization improvement,
for all the design points and various RNN models. Compared to the state-of-the-art ASIC, FPGA, and GPU
implementations, we improve the performance by 2x, 2.8x, and 82x, respectively, considering 64K-MAC. 
Our scheme can also be applied for FPGAs, e.g. BrainWave, besides being targeted for ASIC design.
Finally, we obtain an average utilization of 50\%, for a peak throughput of 30 TFLOPs/s, resulting in 0.32 TFLOPS/Watt 
which is 1.7x and 7.4x more energy-efficient than current ASIC and FPGA implementations, respectively.

%% file: acknowledgement.tex
\section{Acknowledgment}\label{s:Conclusions}
This work has been supported by the CoCoUnit ERC Advanced Grant of the EU’s Horizon 2020 program (grant No 833057), the Spanish State Research Agency (MCIN/AEI) under grant PID2020-113172RB-I00, and the ICREA Academia program.